\documentclass{article} 
\usepackage{iclr2027_conference,times}

\usepackage{amsmath,amsfonts,bm}

\newcommand{\captiona}{{\em (a)}}
\newcommand{\captionb}{{\em (b)}}
\newcommand{\captionc}{{\em (c)}}

\def\eqref#1{equation~\ref{#1}}

\def\1{\bm{1}}

\DeclareMathAlphabet{\mathsfit}{\encodingdefault}{\sfdefault}{m}{sl}
\SetMathAlphabet{\mathsfit}{bold}{\encodingdefault}{\sfdefault}{bx}{n}

\usepackage{hyperref}
\usepackage{url}
\usepackage{booktabs}
\usepackage{graphicx}
\usepackage{multirow}
\usepackage[most]{tcolorbox}
\usepackage{colortbl}
\usepackage{times}
\usepackage{latexsym}
\usepackage{amsmath}
\usepackage{amsfonts}
\usepackage{amssymb}
\usepackage{multirow}
\usepackage{booktabs}
\usepackage{colortbl}
\usepackage{xcolor}
\usepackage{pifont}
\usepackage{subcaption}
\hypersetup{
    hidelinks
}

\usepackage{acro}

\DeclareAcronym{llm}{
  short=LLM,
  long=large language model,
}

\DeclareAcronym{dpo}{
  short=DPO,
  long=Direct Preference Optimization
}

\DeclareAcronym{lora}{
  short=LoRA,
  long=Low-Rank Adaptation
}

\DeclareAcronym{ood}{
  short=OOD,
  long=out-of-distribution
}

\DeclareAcronym{sae}{
  short=SAE,
  short-plural=SAEs,
  long=sparse autoencoder,
  long-plural=sparse autoencoders
}

\DeclareAcronym{mlp}{
  short=MLP,
  long=multilayer perceptron
}

\DeclareAcronym{rope}{
  short=RoPE,
  long=rotary position embedding
}

\DeclareAcronym{icc}{
  short=ICC,
  long=intraclass correlation coefficient
}

\usepackage{xspace}

\newcommand{\LlamaThreeTwoThreeB}{Llama-3.2-3B\xspace}
\newcommand{\LlamaThreeOneEightB}{Llama-3.1-8B\xspace}

\newcommand{\GemmaTwoTwoB}{Gemma-2-2B\xspace}
\newcommand{\GemmaTwoNineB}{Gemma-2-9B\xspace}

\newcommand{\QwenThreeZeroPointSixB}{Qwen3-0.6B\xspace}
\newcommand{\QwenThreeOnePointSevenB}{Qwen3-1.7B\xspace}
\newcommand{\QwenThreeFourB}{Qwen3-4B\xspace}

\newcommand{\GPTFourOMini}{GPT-4o-mini\xspace}
\newcommand{\GeminiFlashLite}{Gemini-3.1-Flash-Lite\xspace}
\newcommand{\ClaudeHaiku}{Claude Haiku 4.5\xspace}

\title{MetaSteer: Context-Conditioned, nonlinear Steering via Attention-Projection Adaptation}

\author{
Mehdi Jafari$^{1,3}$ \qquad
Hao Xue$^{1,2,3}$ \qquad
Flora Salim$^{1,3}$ \\[3pt]
$^{1}$School of Computer Science and Engineering, UNSW Sydney, Australia \\
$^{2}$The Hong Kong University of Science and Technology (Guangzhou), China \\
$^{3}$ARC Centre of Excellence for Automated Decision-Making and Society (ADM+S) \\[3pt]
\texttt{\{mehdi.jafari, flora.salim\}@unsw.edu.au} \qquad
\texttt{haoxue@hkust-gz.edu.cn}
}

\newcommand{\cmark}{\ding{51}}
\newcommand{\xmark}{\ding{55}}
\newcommand{\warnmark}{$\triangle$}

\iclrfinalcopy 
\begin{document}

\maketitle

\begin{abstract}
Steering large language models typically relies on linear, context-independent interventions in activation space, an assumption that recent work has challenged and that can induce an information bottleneck when a fixed representation must encode many behavioral distinctions. We introduce \textit{MetaSteer}, a method that learns nonlinear interventions with context-dependent effects and applies them to attention projection matrices, producing activation effects that vary with the input context by construction and requiring no linear concept-geometry assumption. Framed as preference-based optimization, MetaSteer is trained once on a pooled preference corpus and transferred zero-shot to unseen concepts and out-of-distribution contexts. We find that, despite using low-rank adapters, MetaSteer induces structured, context-dependent changes in hidden-state trajectories while partially preserving aspects of their local trajectory dynamics, including velocity and curvature.  We evaluate MetaSteer on three controlled text-generation benchmarks and three agentic settings across multiple model families and scales. MetaSteer matches or outperforms strong task-specific steering baselines on most aggregate comparisons in the zero-shot regime. Across the evaluated settings, stronger text-generation steering is associated with stronger agentic steering performance. We further discuss geometric trajectory effects, capability retention, and safety considerations raised by transferable steering.
\end{abstract}

\section{Introduction}

Steering in \acp{llm} refers to applying interventions to the activation space to control their generations \citep{wu2025axbench}. Such an intervention can be characterized by three design choices: the geometric assumption underpinning it -- traditionally linear \citep{xiong2024everything, bereska2024mechanistic, saglam2025large, panickssery2023steering, turner2023steering}, or nonlinear and context-dependent in light of evidence that concepts lie on curved, anisotropic manifolds \citep{engels2025not, modell2025origins, nguyen2026beyond, zhao2026odesteer, mishra2026steered} -- the location within the LLM's architecture where it is applied, namely the residual stream \citep{nguyen2025multi, hsu2026contextual}, MLP components \citep{yu2026wasd, yan2026spurious}, or attention heads \citep{genadi2026sycophancy, luo2026don} -- and the procedure used to compute it, whether heuristic \citep{chalnev2024improving}, causal-effect prediction \citep{arad2025saes, cho2025corrsteer, soo2025interpretable}, preference-based cloning \citep{raina2025d}, or end-to-end learned vectors \citep{sun2025hypersteer}. 
The interplay among these choices can be formulated as a \emph{general dual-objective optimization} problem \citep{aravindan2026opium, nguyen2026minimizing, luo2026learning}, at the heart of which lies the fundamental challenge common to every steering problem: balancing steerability ($\mathbb{S}$) with preservation of the model's general utility ($\mathbb{U}$).

Although concept- or task-specific steering methods for LLMs exist, a general approach that balances steering effectiveness with preservation of the model's capabilities has yet to be established. Such an approach should avoid overly restrictive assumptions about the geometry of concepts and semantic space so that it can generalize to unseen tasks and concepts; intervene at a location that is both effective and computationally tractable; and remain minimally invasive, without degrading the model's general capabilities. This raises a concrete question: can we learn a single intervention that adapts its effect to the current context, transfers to unseen concepts and tasks, and preserves the model's general capabilities?

We answer this question with \textit{MetaSteer}, a preference-trained, nonlinear intervention applied to the attention projection matrices. We choose this site because it offers a computationally tractable way to modify the model's internal computation \citep{luo2026don} while allowing its effects to propagate through residual connections, feed-forward sublayers, and subsequent attention blocks. The intervention therefore need not remain confined to a fixed activation-space shift. Because attention is itself a function of the context, MetaSteer's adapter weights, although fixed after training, induce context-dependent effects on the residual stream trajectory: the same parameters can produce different activation shifts depending on the context. MetaSteer is trained once on a pooled preference corpus spanning diverse instructions and steering concepts, then deployed with fixed parameters for zero-shot transfer to unseen concepts and out-of-distribution contexts while largely preserving the model's general capabilities.

We measure transfer on three text-generation benchmarks -- AxBench \citep{wu2025axbench}, CLaS-Bench \citep{gurgurov2026clas}, and PersonalityBench \citep{deng2025neuron} -- and test whether the same intervention transfers to agentic settings through SocialEval \citep{zhou2025socialeval} and the Dictator and Ultimatum Games \citep{mozikov2024eai}. MetaSteer matches or exceeds strong task-specific baselines on aggregate steering scores in most tested settings\footnote{All models are instruction-tuned unless explicitly identified as base models.}.

\begin{figure*}[t]
    \centering

    \begin{minipage}[t]{0.21\textwidth}
        \centering
        \includegraphics[width=\linewidth]{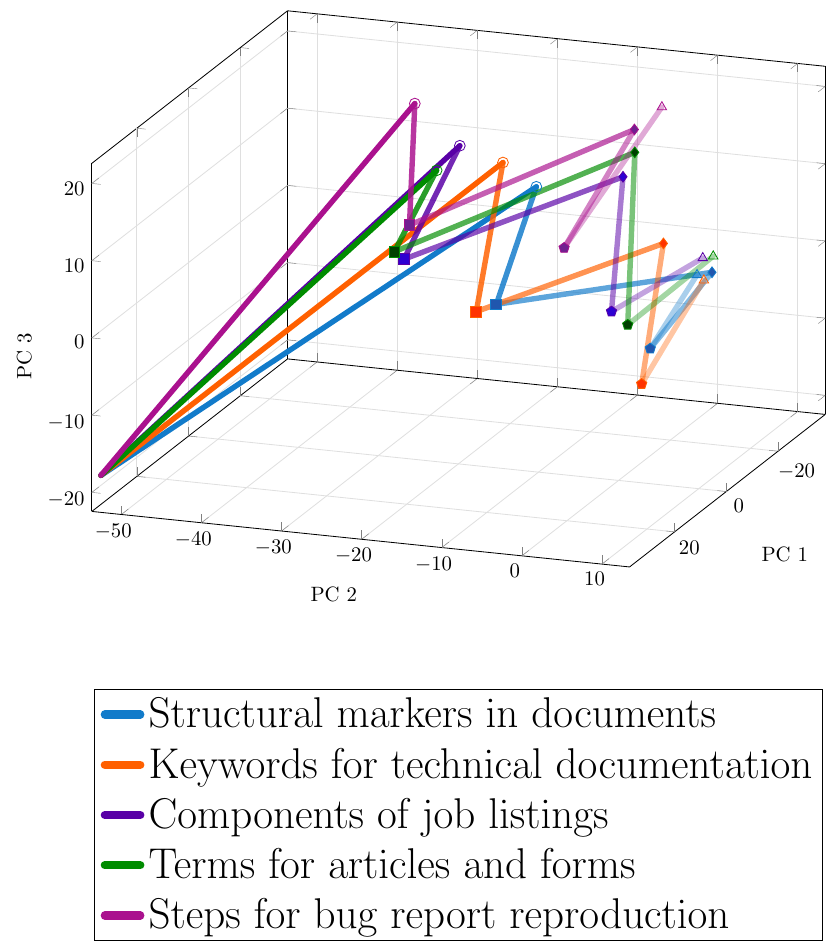}
        \captiona
    \end{minipage}
    \hfill
    \begin{minipage}[t]{0.21\textwidth}
        \centering
        \includegraphics[width=\linewidth]{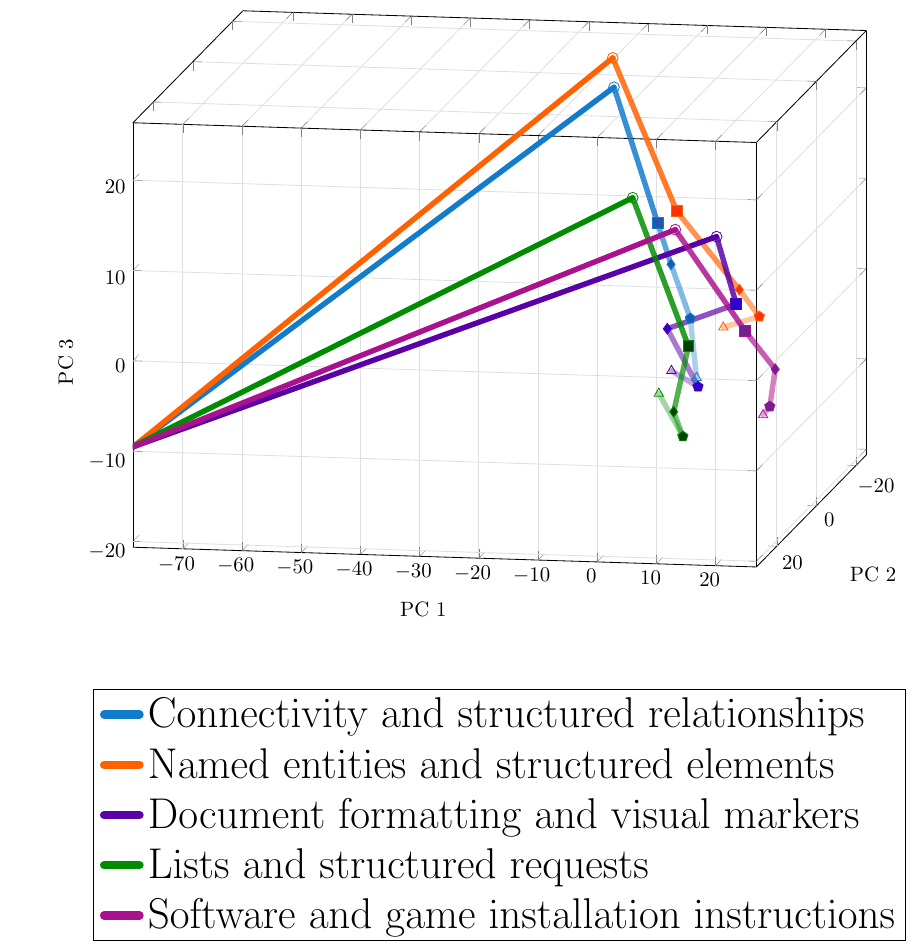}
        \captionb
    \end{minipage}
    \hfill
    \begin{minipage}[t]{0.55\textwidth}
        \centering
        \includegraphics[width=\linewidth]{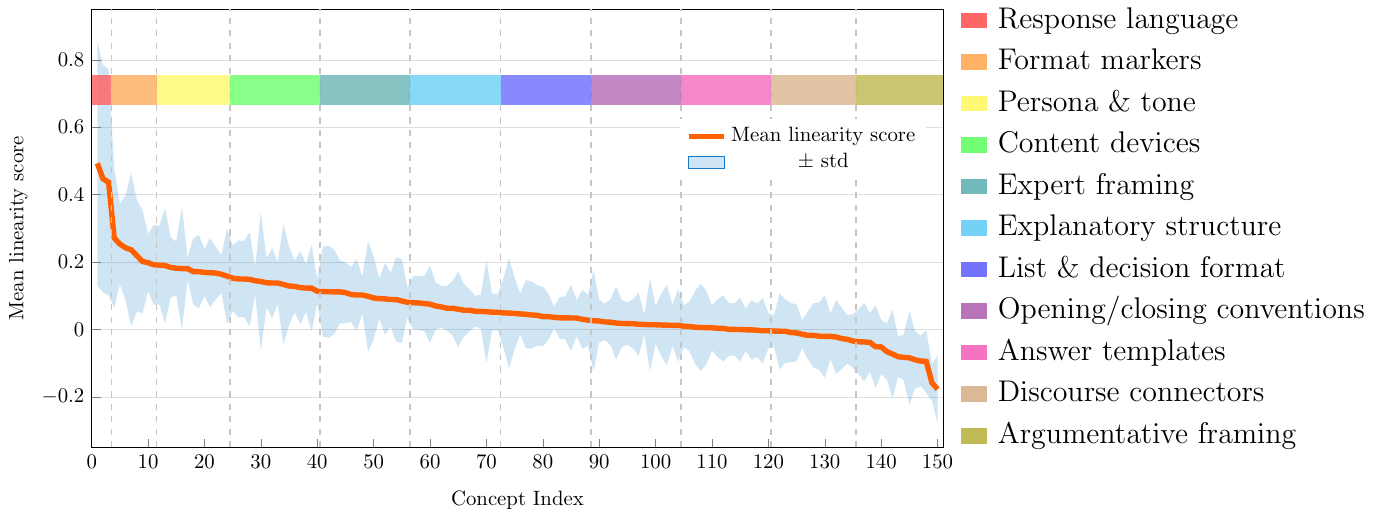}
        \captionc
    \end{minipage}

    \caption{\textbf{Hidden-state trajectories and steering alignment.}
    \textbf{\captiona}--\textbf{\captionb} show PCA projections of sentence-level
    hidden states under different steering concepts. Steering shifts trajectory
    positions while retaining aspects of velocity and curvature.\protect\footnotemark[2]
    \textbf{\captionc} reports cosine alignment between MetaSteer and CAA
    displacements across semantic concept groups; higher values indicate closer
    agreement with linear steering.
    \protect\footnotemark[3]
    }

    \label{fig:three_trajectory_plots}
\end{figure*}

The geometry of steered hidden-state trajectories is also analyzed through
their position, velocity, and curvature. An emerging pattern from this
analysis, illustrated by representative examples in
Fig.~\ref{fig:three_trajectory_plots}(a)--(c), is that steering can displace
trajectory position while preserving some aspects of how the trajectory
evolves, with the degree of linearity varying across steering concepts.
Capability retention is assessed separately, alongside a discussion of the
safety implications of transferable steering.

\footnotetext[2] {Further conceptual details are provided in Appendix~\ref{app:cross_model_similarity}, with additional information about these two instances given in Appendix~\ref{app:steering_geometric_transformation}. }
\footnotetext[3]{%
See Appendix~\ref{app:concept_linearity} for experimental details and
Appendix~\ref{app:concept_inventory} for the complete list of concept
clusters.
}

Our contributions are summarized as follows:

\begin{itemize}
    \item \textbf{A nonlinear attention-based intervention.}
    \textit{MetaSteer} is a preference-trained, low-rank intervention applied
    jointly to the query, key, value, and output attention projections, with
    an accompanying mathematical formalization.

    \item \textbf{Zero-shot transfer across concepts and settings.}
    A single context-conditioned intervention per backbone, trained on pooled
    concept--preference data and fixed after training, transfers to held-out
    concepts, external benchmarks and agentic settings without further adaptation.

    \item \textbf{Geometric characterization of steering.}
    The hidden-state trajectories induced by steering are analyzed through
    their position, velocity, and curvature. The results provide evidence
    that steering can displace trajectory position while preserving some
    aspects of trajectory evolution, with the degree of linearity varying
    across steering concepts.
\end{itemize}

\section{Related Work}
\label{sec:related_work}

Steering methods can be organized around three design questions: what
geometric structure is assumed for concepts in the model's semantic space,
where the intervention is applied, and how the intervention is computed
\footnote{A detailed discussion is provided in
Appendix~\ref{app:extended_related_work}.}.

\textbf{Geometric assumption.} Much prior work assumes concepts are represented approximately linearly in activation space, motivating direction- or vector-difference-based interventions \citep{xiong2024everything,bereska2024mechanistic,saglam2025large,panickssery2023steering,turner2023steering} that compose and transfer across concepts \citep{karvonen2024emergent,nguyen2025multi}. Recent work instead shows concepts may occupy curved, anisotropic manifolds, motivating nonlinear, context-dependent interventions \citep{engels2025not,modell2025origins,nguyen2026beyond,zhao2026odesteer,mishra2026steered}. \emph{MetaSteer} imposes no such restrictive geometric assumption on the structure of semantic space.

\textbf{Intervention site.} Interventions may act at the input level via
prompts, personas, or reasoning instructions
\citep{kong-etal-2024-better,miehling2025evaluating,
park2025iclr,wei2022chain,wu2025axbench}, thereby modifying the conditioning context
rather than applying an explicit intervention vector; at the parameter level
via adapters, prompt or prefix tuning, LoRA, or representation fine-tuning
\citep{pmlr-v97-houlsby19a,lester2021power,li2021prefix,hu2022lora,
trung-etal-2024-reft}; or within the forward pass---most commonly through
the residual stream \citep{nguyen2025multi,hsu2026contextual} or
sparse-autoencoder features \citep{yu2026wasd,yan2026spurious}, with recent
work also targeting MLP components and attention heads
\citep{genadi2026sycophancy,luo2026don}. The closest work to ours is
\citet{luo2026don}, which intervenes only on the query projection; by
contrast, \textit{MetaSteer} intervenes jointly on all four attention
projection matrices---the query, key, value, and output projections.

\textbf{Learning procedure.} Interventions have been constructed via heuristic or contrastive procedures \citep{panickssery2023steering, chalnev2024improving}; objectives balancing steering effectiveness against utility preservation \citep{aravindan2026opium,nguyen2026minimizing,luo2026learning}; learned causal-effect estimators and steering operators \citep{arad2025saes,cho2025corrsteer,soo2025interpretable,sun2025hypersteer}; and preference-based activation-space behavior cloning \citep{raina2025d}. MetaSteer is closest in learning approach to \citep{raina2025d}, but avoids the information bottleneck they identify by learning nonlinear, context-dependent interventions rather than a single low-dimensional vector. Unlike \citep{sun2025hypersteer}, which extracts steering directions using a separate, architecturally identical model, MetaSteer uses the same model.

\section{Methodology}

We formulate steering as preference optimization of a shared, low-rank
intervention. We first describe DPO as a direction-specific steering baseline,
then introduce an intervention on the four attention projections. We
subsequently characterize the geometry of the resulting activation updates and
describe the preference data and training objective.

\subsection{DPO as Direction-Specific Steering}
\label{sec:background}

DPO increases the likelihood of a preferred completion $y^+$ relative to a
dispreferred completion $y^-$ for a prompt $x$. For a single next-token
comparison at a shared prompt $x$, the log-probability gap between the
preferred and dispreferred token is linear in the model's final hidden
state $\mathbf{h}(x)$:
\begin{equation}
\log\pi(y^+\mid x) - \log\pi(y^-\mid x)
=
\langle \mathbf{h}(x), \mathbf{v}\rangle,
\qquad
\mathbf{v} := \mathbf{e}_{y^+} - \mathbf{e}_{y^-},
\label{eq:logit-gap}
\end{equation}
an identity that follows directly from the softmax output parameterization
and is derived in full in Appendix~\ref{app:logit-gap} \citep{raina2025d}.
For a fixed candidate-token pair $(y^+,y^-)$, the gradient of this
single-step gap with respect to the hidden state is parallel to
$\mathbf{v}$. This identity alone does not imply a common direction across
examples when the candidate-token pair varies; the approximately fixed
$\mathbf{d}^{\star}$ in Eq.~\ref{eq:direction-steering} additionally relies
on the local approximation of \citet{raina2025d}. Under a local
linearization, the resulting activation update can therefore be
approximated as
\begin{equation}
\mathbf{h}_{\mathrm{DPO}}(x)
\approx
\mathbf{h}_0(x)+\alpha\,\mathbf{d}^{\star},
\label{eq:direction-steering}
\end{equation}
where $\mathbf{d}^{\star}$ is approximately fixed across examples
\citep{raina2025d}. Equation~\ref{eq:logit-gap} is stated here for a
single next-token comparison; for full multi-token completions, it holds
exactly at the position where the two completions first diverge, with the
remainder of the sequence gap following an ordinary autoregressive
expansion rather than a single fixed direction (Appendix~\ref{app:logit-gap}).
This direction-specific formulation is efficient, but it cannot represent
arbitrary context-dependent behavior and relies on an approximately linear
representation of the target concept
\citep{engels2025not,modell2025origins}.

\subsection{Context-Conditioned Attention Intervention}
\label{sec:mechanism}

Rather than adding a fixed vector to the residual stream, we apply low-rank
adapters to all four attention projection matrices,
$f\in\{q,k,v,o\}$, at each layer:
\begin{equation}
\widetilde{\mathbf{W}}_f
=
\mathbf{W}_f+\Delta\mathbf{W}_f
=
\mathbf{W}_f+\lambda_f\mathbf{A}_f\mathbf{B}_f^{\top},
\qquad r\ll d,
\label{eq:lora-adapters}
\end{equation}
where $\mathbf{A}_f$ and $\mathbf{B}_f$ are trainable low-rank factors and
$\lambda_f$ controls the intervention intensity.

The direct update remains rank-constrained: for each projection,
$\lambda_f\mathbf{z}_i\mathbf{A}_f\mathbf{B}_f^\top$ lies in an
at-most-$r$-dimensional subspace. Context changes its coefficients within
this subspace, while attention and composition across layers yield a richer
end-to-end effect.

Let $\mathbf{z}_i$ denote the normalized activation at token position $i$.
The baseline and adapted query, key, and value representations are
\[
\mathbf{q}_i=\mathbf{z}_i\mathbf{W}_q,\qquad
\mathbf{k}_j=\mathbf{z}_j\mathbf{W}_k,\qquad
\mathbf{v}_j=\mathbf{z}_j\mathbf{W}_v,
\]
\[
\widetilde{\mathbf{q}}_i=\mathbf{z}_i\widetilde{\mathbf{W}}_q,\qquad
\widetilde{\mathbf{k}}_j=\mathbf{z}_j\widetilde{\mathbf{W}}_k,\qquad
\widetilde{\mathbf{v}}_j=\mathbf{z}_j\widetilde{\mathbf{W}}_v,
\]
with $\Delta\mathbf{q}_i:=\widetilde{\mathbf{q}}_i-\mathbf{q}_i$ and
$\Delta\mathbf{k}_j:=\widetilde{\mathbf{k}}_j-\mathbf{k}_j$ following from
Eq.~\ref{eq:lora-adapters}. For a token pair $(i,j)$, the (unperturbed)
scaled dot-product attention logit is $s_{ij}=\langle\mathbf{q}_i,\mathbf{k}_j\rangle/\sqrt{d}$,
and the adapted logit is $\widetilde{s}_{ij}=\langle\widetilde{\mathbf{q}}_i,\widetilde{\mathbf{k}}_j\rangle/\sqrt{d}$,
which we write as
\begin{equation}
\widetilde{s}_{ij}
=
s_{ij}+\delta_{ij}(\mathbf{z}_i,\mathbf{z}_j),
\label{eq:attention-perturbation}
\end{equation}
where, expanding
$\langle\widetilde{\mathbf{q}}_i,\widetilde{\mathbf{k}}_j\rangle
=\langle\mathbf{q}_i+\Delta\mathbf{q}_i,\mathbf{k}_j+\Delta\mathbf{k}_j\rangle$
and cancelling the shared $\langle\mathbf{q}_i,\mathbf{k}_j\rangle/\sqrt{d}=s_{ij}$
term, the perturbation is exactly the sum of the three cross terms induced
by the low-rank update:
\begin{equation}
\delta_{ij}(\mathbf{z}_i,\mathbf{z}_j)
=
\frac{1}{\sqrt{d}}
\Big[
\langle\mathbf{q}_i,\Delta\mathbf{k}_j\rangle
+
\langle\Delta\mathbf{q}_i,\mathbf{k}_j\rangle
+
\langle\Delta\mathbf{q}_i,\Delta\mathbf{k}_j\rangle
\Big].
\label{eq:delta-ij}
\end{equation}
$\delta_{ij}$ depends on both token representations because the adapters
modify the query and key projections. For generic adapters,
\begin{equation}
\frac{\partial\delta_{ij}}{\partial\mathbf{z}_i}\neq 0,
\qquad
\frac{\partial\delta_{ij}}{\partial\mathbf{z}_j}\neq 0.
\label{eq:nonzero-grad}
\end{equation}
Thus, although the parameter update is fixed after training, its induced
attention change depends on the surrounding input context. The full
perturbation is derived in Appendix~\ref{app:proofs}.

For notational simplicity, Equations~\ref{eq:attention-perturbation}
and~\ref{eq:delta-ij} are stated for a single attention head with no
positional rotation; the general form for multi-head, grouped-query
attention under RoPE is derived in
Appendix~\ref{app:proofs-multihead} and reduces to those equations exactly
when $H=H_{kv}=1$ and the rotation is the identity. In the general case,
with $H$ query heads, $H_{kv}\leq H$ key/value heads (query head $h$
sharing key/value head $\kappa(h)$ under grouped-query attention), and
per-head outputs $\widetilde{\mathbf{a}}_i^{(h)}$, the output projection is
applied \emph{after} concatenating head outputs rather than to each value
vector independently:
\begin{equation}
\widetilde{\mathbf{a}}_i
=
\operatorname{Concat}_{h=1}^{H}\!\big(\widetilde{\mathbf{a}}_i^{(h)}\big)\,
\widetilde{\mathbf{W}}_o
=
\sum_{h=1}^{H}
\sum_j
\operatorname{softmax}_j\!\big(\widetilde{s}_{ij}^{(h)}\big)
\left(\widetilde{\mathbf{v}}_j^{(\kappa(h))}\widetilde{\mathbf{W}}_o^{(h)}\right),
\label{eq:adapted-attention}
\end{equation}
where $\widetilde{\mathbf{W}}_o^{(h)}$ is the row-block of
$\widetilde{\mathbf{W}}_o$ corresponding to head $h$, and
$\widetilde{s}_{ij}^{(h)}$ is that head's own adapted attention logit
(Appendix~\ref{app:proofs-multihead}). Because the softmax weights and
projected values depend on the current token representations, the induced
update $\Delta\mathbf{a}_i=\widetilde{\mathbf{a}}_i-\mathbf{a}_i$ is
generally nonlinear:
\begin{equation}
\frac{\partial^2\Delta\mathbf{a}_i}
{\partial\mathbf{z}_i^2}
\not\equiv 0.
\label{eq:nonlinear}
\end{equation}
This provides a context-conditioned alternative to adding a fixed activation
direction.

\subsection{Geometry of Steering Effects}
\label{sec:steering_geometry}

The preceding parameterization produces an activation displacement that can
vary with both the instruction and the steering concept. We write this
displacement as
\begin{equation}
\mathbf{h}_{\Theta}(x,c)
\approx
\mathbf{h}_0(x)+\alpha(x,c)\mathbf{d}(x,c),
\qquad
\mathbf{d}(x,c)\in\mathbb{R}^{d},
\label{eq:concept-displacement}
\end{equation}
where $\Theta$ denotes the adapter parameters.

We test whether this context-dependent displacement nonetheless collapses,
concept by concept, onto a single fixed linear direction, or instead departs
from one. For each steering concept $c$, we compare its mean MetaSteer
displacement $\mathbf{d}^{\mathrm{MS}}_c := \mathbb{E}_{x}[\mathbf{d}(x,c)]$
against CAA direction
$\mathbf{v}^{\mathrm{CAA}}_c$ estimated for the same concept, via their cosine
similarity
\begin{equation}
s_c = \cos\!\left(\mathbf{v}^{\mathrm{CAA}}_c,\ \mathbf{d}^{\mathrm{MS}}_c\right).
\label{eq:caa-alignment}
\end{equation}
A value of $s_c$ near one indicates alignment with the CAA direction, while a
low value indicates disagreement with this particular linear baseline. This is
an alignment diagnostic; it neither rules out another fixed linear direction
nor estimates the intrinsic dimensionality of the hidden-state trajectory.
We report $s_c$ for a broad concept inventory across multiple models in
Section~\ref{sec:geometry} and Appendix~\ref{app:concept_linearity}
(Figure~\ref{fig:three_trajectory_plots}\captionc).

\subsection{Preference Data}
\label{sec:data}

We construct a pooled preference dataset from Concept16K and Concept16K-v2
\citep{wu2025axbench}. For each steering concept $c$ and instruction $x$, the
dataset provides preferred and dispreferred responses. We form DPO tuples
\begin{equation}
(x,c,y^+,y^-),
\qquad
y^+\in\mathcal{D}_c^+,\quad
y^-\in\mathcal{D}_c^-,
\label{eq:tuples}
\end{equation}
and pool them across concepts:
\begin{equation}
\mathcal{D}
=
\bigcup_{c\in\mathcal{C}}
\{(x,c,y^+,y^-)\}.
\label{eq:data-pool}
\end{equation}
The internal split is concept-disjoint but shares instructions across splits;
transfer to external instruction distributions is evaluated only on the
external benchmarks. Dataset construction and diversity statistics are
provided in Appendix~\ref{app:axbench_dataset}.

\subsection{Preference Optimization Objective}
\label{sec:objective}

Let $\pi_{\Theta}$ denote the base model with the attention adapters and let
$\pi_{\mathrm{ref}}$ be the frozen reference model. We define
\begin{equation}
\rho_{\Theta}(y\mid x,c)
=
\log\pi_{\Theta}(y\mid x,c)
-
\log\pi_{\mathrm{ref}}(y\mid x,c).
\label{eq:rho-conditioned}
\end{equation}

We optimize the adapter factors $\Theta=\{\mathbf{A}_f,\mathbf{B}_f\}$,
conditioned on the instruction and steering concept, using the standard
DPO objective with fixed $\lambda_f=1$ (Appendix~\ref{app:lora_details}):
\begin{equation}
\mathcal{L}_{\mathrm{DPO}}(\Theta)
=
-\mathbb{E}_{(x,c,y^+,y^-)\sim\mathcal{D}}
\left[
\log\sigma\left(
\beta\rho_{\Theta}(y^+\mid x,c)
-
\beta\rho_{\Theta}(y^-\mid x,c)
\right)
\right].
\label{eq:conditioned-dpo}
\end{equation}

The base model parameters remain frozen; only the low-rank adapter factors
$\{\mathbf{A}_f,\mathbf{B}_f\}$ are updated. The intensity scalars $\lambda_f$
are fixed hyperparameters (Appendix~\ref{app:lora_details}).

\begin{table*}[t]
\centering
\small
\setlength{\tabcolsep}{5.1pt}
\renewcommand{\arraystretch}{1.05}

\caption{
Zero-shot transfer results on AxBench, CLaS-Bench, and PersonalityBench across
the evaluated LLaMA, Gemma, and Qwen model scales. Aggregate-score columns
summarize the metrics described in Section~\ref{sec:zero-shots-text}. Bold entries
indicate the larger aggregate score for each model.
}
\label{tab:transfer-results}

\begin{tabular}{ll cccc ccc ccc}
\toprule
\multirow{2}{*}{\textbf{Model}}
& \multirow{2}{*}{\textbf{Method}}
& \multicolumn{4}{c}{\textbf{AxBench}}
& \multicolumn{3}{c}{\textbf{CLaS-Bench}}
& \multicolumn{3}{c}{\textbf{PersonalityBench}} \\

\cmidrule(lr){3-6}
\cmidrule(lr){7-9}
\cmidrule(lr){10-12}

& & \textbf{C} & \textbf{I} & \textbf{F} & \textbf{HM}
& \textbf{L} & \textbf{R} & \textbf{HM}
& \textbf{P} & \textbf{F} & \textbf{AM} \\
\midrule

\multirow{2}{*}{\LlamaThreeTwoThreeB}
& \textsc{Best Ref}
& 0.63 & 1.78 & 1.07 & 0.97
& 71.61 & 64.43 & 67.83
& 4.87 & 5.00 & \textbf{4.94} \\

& \textit{MetaSteer}
& 1.65 & 1.75 & 1.49 & \textbf{1.62}
& 79.64 & 71.59 & \textbf{75.40}
& 4.86 & 5.00 & 4.93 \\
\midrule

\multirow{2}{*}{\LlamaThreeOneEightB}
& \textsc{Best Ref}
& 1.22 & 1.80 & 1.10 & 1.32
& 83.50 & 85.50 & 84.50
& 4.89 & 5.00 & \textbf{4.95} \\

& \textit{MetaSteer}
& 1.68 & 1.82 & 1.27 & \textbf{1.55}
& 87.07 & 87.25 & \textbf{87.16}
& 4.87 & 4.99 & 4.93 \\
\midrule

\multirow{2}{*}{\GemmaTwoTwoB}
& \textsc{Best Ref}
& 0.74 & 1.77 & 1.06 & 1.05
& 83.26 & 79.03 & 81.09
& 4.84 & 5.00 & 4.92 \\

& \textit{MetaSteer}
& 1.42 & 1.47 & 1.67 & \textbf{1.51}
& 94.23 & 78.71 & \textbf{85.78}
& 4.91 & 5.00 & \textbf{4.95} \\
\midrule

\multirow{2}{*}{\GemmaTwoNineB}
& \textsc{Best Ref}
& 1.03 & 1.80 & 1.12 & 1.24
& 97.81 & 90.43 & 93.98
& 4.86 & 5.00 & 4.93 \\

& \textit{MetaSteer}
& 1.42 & 1.58 & 1.58 & \textbf{1.52}
& 99.49 & 90.30 & \textbf{94.67}
& 4.96 & 5.00 & \textbf{4.98} \\
\midrule

\multirow{2}{*}{\QwenThreeOnePointSevenB}
& \textsc{Best Ref}
& 1.02 & 1.85 & 1.41 & 1.35
& 60.09 & 65.40 & 62.63
& 4.91 & 4.99 & \textbf{4.95} \\

& \textit{MetaSteer}
& 1.45 & 1.72 & 1.56 & \textbf{1.57}
& 73.43 & 71.65 & \textbf{72.53}
& 4.84 & 5.00 & 4.92 \\
\midrule

\multirow{2}{*}{\QwenThreeFourB}
& \textsc{Best Ref}
& 1.40 & 1.89 & 1.40 & 1.53
& 83.26 & 77.65 & 80.36
& 4.89 & 5.00 & 4.94 \\

& \textit{MetaSteer}
& 1.58 & 1.74 & 1.46 & \textbf{1.59}
& 85.52 & 77.99 & \textbf{81.58}
& 4.91 & 5.00 & \textbf{4.95} \\

\bottomrule
\end{tabular}
\end{table*}

\section{Zero-Shot Transfer to Text-Generation Benchmarks}
\label{sec:zero-shots-text}

We evaluate whether \textit{MetaSteer}, trained once per backbone on the pooled
preference corpus $\mathcal{D}$, transfers to held-out concepts and external
benchmark instruction distributions. No per-benchmark
fine-tuning or task-specific hyperparameter search is performed.

\paragraph{Benchmarks.}

\noindent\textbf{AxBench.} \citep{wu2025axbench} evaluates instruction following under
concept-steering constraints across information-seeking, mathematical, and
programming queries ($\sim$32K test examples). Its metrics are concept score
(C), instruction score (I), and fluency score (F), combined using their
harmonic mean (HM). \textbf{CLaS-Bench.} \citep{gurgurov2026clas} evaluates language steering in
question-answering settings ($\sim$72K test examples). Its metrics are
language forcing (L) and output relevance (R), combined using their harmonic
mean (HM). \textbf{PersonalityBench.}
PersonalityBench \citep{deng2025neuron} evaluates elicitation of the Big Five
personality traits through open-ended long responses ($\sim$500 test examples). Its
metrics are personality score (P) and fluency score (F), whose arithmetic mean
(AM) is reported as the aggregate score.

\paragraph{Reference results and matched baselines.}
We use two complementary comparisons for the text-generation tasks. The
\textsc{Best Ref} row denotes the best-performing method reported for each
benchmark: \emph{Prompt} for AxBench, $\varepsilon$-base-$I$ for CLaS-Bench,
and \emph{P2} for PersonalityBench. We implemented each method and evaluated
it using the corresponding benchmark's original framework and scoring
procedure. \textit{MetaSteer} was then evaluated on the same tasks under the
same framework, enabling a direct comparison with the strongest task-specific
reference available for each benchmark.

We additionally evaluate SKOP, the closest methodological comparator to
\textit{MetaSteer}, in a matched head-to-head comparison following SKOP's
reported evaluation procedure. On \LlamaThreeOneEightB, \textit{MetaSteer} outperforms SKOP by
2\% in utility ($\mathbb{U}$) and 0.6 points in steering effectiveness
($\mathbb{S}$). HyperSteer is discussed as a related
transferable-steering method, but its reported AxBench performance is below
the task-specific reference implemented here. D-Steer is also conceptually
related; however, its fixed representation introduces an information
bottleneck that limits its ability to represent fine-grained concept
variation. Results for three model families at two parameter scales are reported in
Table~\ref{tab:transfer-results}.

\paragraph{Results.}
\textit{MetaSteer} improves aggregate steering scores over the reported
baselines on AxBench and CLaS-Bench across settings. On
PersonalityBench, it remains competitive, with only small differences in
aggregate scores. Together with the zero-shot evaluation and the geometric
analysis in Section~\ref{sec:geometry}, these results support structured and
transferable steering effects rather than collapse to a single direction or
concept-specific memorization.

\begin{table*}[t]
\centering
\small
\setlength{\tabcolsep}{3pt}
\renewcommand{\arraystretch}{1.05}
\caption{
Emotion-steering results for the Dictator and Ultimatum Games across three
model families at two parameter scales, with GPT-4o and Human reference rows
(scoring defined in Section~\ref{sec:zero-shots-agent}). \textbf{Bold} indicates the
best Simulation Score per model.
}
\label{tab:emotion-results}

\begin{tabular}{ll ccc ccc ccc ccc ccc c}
\toprule

\multirow{2}{*}{\textbf{Agent}}
& \multirow{2}{*}{\textbf{Method}}
& \multicolumn{3}{c}{\textbf{Anger}}
& \multicolumn{3}{c}{\textbf{Disgust}}
& \multicolumn{3}{c}{\textbf{Fear}}
& \multicolumn{3}{c}{\textbf{Happiness}}
& \multicolumn{3}{c}{\textbf{Sadness}}
& \multirow{2}{*}{\shortstack{\textbf{Simulation}\\\textbf{Score}}} \\

\cmidrule(lr){3-5}
\cmidrule(lr){6-8}
\cmidrule(lr){9-11}
\cmidrule(lr){12-14}
\cmidrule(lr){15-17}

& & \textbf{D} & \textbf{UP} & \textbf{UR}
& \textbf{D} & \textbf{UP} & \textbf{UR}
& \textbf{D} & \textbf{UP} & \textbf{UR}
& \textbf{D} & \textbf{UP} & \textbf{UR}
& \textbf{D} & \textbf{UP} & \textbf{UR}
& \\

\midrule

\multicolumn{2}{l}{Human}
& $\uparrow$ & $\uparrow$ & $\downarrow$
& $\downarrow$ & $\downarrow$ & $\uparrow$
& $\uparrow$ & $\uparrow$ & $\uparrow$
& $\downarrow$ & $\downarrow$ & $\downarrow$
& $\uparrow$ & $\uparrow$ & $\downarrow$
& -- \\

\multicolumn{2}{l}{GPT-4o}
& \xmark & \xmark & \cmark
& \xmark & \warnmark & \xmark
& \cmark & \cmark & \cmark
& \xmark & \xmark & \xmark
& \cmark & \cmark & \cmark
& 15/30 \\

\midrule

\multirow{2}{*}{\LlamaThreeTwoThreeB}
& \textsc{EAI-CP}
& \xmark & \xmark & \cmark
& \xmark & \cmark & \xmark
& \xmark & \xmark & \xmark
& \cmark & \cmark & \cmark
& \xmark & \xmark & \warnmark
& 11/30 \\

& \textit{MetaSteer}
& \xmark & \xmark & \warnmark
& \cmark & \cmark & \xmark
& \xmark & \xmark & \xmark
& \cmark & \cmark & \cmark
& \cmark & \xmark & \cmark
& \textbf{15/30} \\

\midrule

\multirow{2}{*}{\LlamaThreeOneEightB}
& \textsc{EAI-CP}
& \xmark & \xmark & \cmark
& \xmark & \xmark & \xmark
& \xmark & \cmark & \xmark
& \cmark & \xmark & \cmark
& \cmark & \cmark & \cmark
& 14/30 \\

& \textit{MetaSteer}
& \cmark & \xmark & \cmark
& \xmark & \xmark & \xmark
& \cmark & \cmark & \xmark
& \cmark & \cmark & \cmark
& \cmark & \cmark & \cmark
& \textbf{20/30} \\

\midrule

\multirow{2}{*}{\GemmaTwoTwoB}
& \textsc{EAI-CP}
& \xmark & \cmark & \xmark
& \xmark & \xmark & \xmark
& \cmark & \cmark & \xmark
& \cmark & \xmark & \cmark
& \cmark & \cmark & \cmark
& 16/30 \\

& \textit{MetaSteer}
& \cmark & \cmark & \cmark
& \xmark & \xmark & \xmark
& \cmark & \cmark & \xmark
& \cmark & \cmark & \cmark
& \cmark & \cmark & \cmark
& \textbf{22/30} \\

\midrule

\multirow{2}{*}{\GemmaTwoNineB}
& \textsc{EAI-CP}
& \xmark & \cmark & \xmark
& \cmark & \cmark & \xmark
& \xmark & \cmark & \cmark
& \cmark & \xmark & \xmark
& \cmark & \cmark & \xmark
& 16/30 \\

& \textit{MetaSteer}
& \xmark & \xmark & \cmark
& \xmark & \cmark & \xmark
& \cmark & \xmark & \xmark
& \cmark & \cmark & \warnmark
& \cmark & \cmark & \cmark
& \textbf{17/30} \\

\midrule

\multirow{2}{*}{\QwenThreeOnePointSevenB}
& \textsc{EAI-CP}
& \xmark & \xmark & \cmark
& \cmark & \cmark & \warnmark
& \cmark & \xmark & \warnmark
& \cmark & \cmark & \cmark
& \xmark & \xmark & \cmark
& \textbf{18/30} \\

& \textit{MetaSteer}
& \xmark & \cmark & \xmark
& \cmark & \xmark & \xmark
& \xmark & \warnmark & \warnmark
& \cmark & \cmark & \cmark
& \cmark & \warnmark & \warnmark
& 16/30 \\

\midrule

\multirow{2}{*}{\QwenThreeFourB}
& \textsc{EAI-CP}
& \xmark & \cmark & \warnmark
& \cmark & \xmark & \xmark
& \xmark & \cmark & \warnmark
& \cmark & \xmark & \warnmark
& \cmark & \cmark & \warnmark
& 16/30 \\

& \textit{MetaSteer}
& \xmark & \xmark & \cmark
& \cmark & \cmark & \xmark
& \xmark & \xmark & \warnmark
& \cmark & \cmark & \warnmark
& \cmark & \cmark & \warnmark
& \textbf{17/30} \\

\bottomrule
\end{tabular}
\end{table*}

\section{Beyond Text Generation: Agentic Transfer}
\label{sec:zero-shots-agent}

Transfer on text-generation benchmarks does not necessarily imply transfer to
interactive decision-making. We therefore evaluate whether the same fixed
steering intervention can influence agentic behavior across three settings --
the Dictator Game, the Ultimatum Game, and SocialEval -- while preserving the
capability to track task dynamics and produce valid, in-context actions.

\paragraph{Dictator and Ultimatum Games.}
In the Dictator Game, an allocator divides a fixed amount of money between
itself and a passive recipient. In the Ultimatum Game, a proposer divides a
fixed amount between itself and a responder, who accepts or rejects the
offer; we evaluate both roles. In both games, the agent is steered toward one
of five emotions -- anger, disgust, fear, happiness, or sadness -- and we
compare the resulting change in behavior (the allocator's giving, the
proposer's offer size, and the responder's acceptance rate) against the
human ground-truth direction ($\uparrow,\downarrow$) reported for each
emotion by \citet{mozikov2024eai}.

\paragraph{SocialEval.}
SocialEval \citep{zhou2025socialeval} presents the agent with branching
social scenarios: at each decision point, the agent selects an action that
determines the subsequent trajectory, and each terminal node is annotated
with the social trait it expresses. The agent is steered toward one of three
traits -- proself, antisocial, or prosocial -- and we report the resulting
Goal Achievement Score (GAE) at the macro level (averaged across trait
categories) and the micro level (averaged across all trajectories).

\paragraph{Reference points and metrics.}
For SocialEval, \textsc{SocialEval-En} (\textsc{Seval-En} in
Table~\ref{tab:gae-results}) is the reference configuration, evaluated on the
same English scenarios, trait categories, and macro/micro GAE metrics as
\textit{MetaSteer}. For the Dictator and Ultimatum Games, the \textsc{EAI-CP}
(Co-player) condition of \citet{mozikov2024eai} is the reference, evaluated
under the same emotion--role conditions. In Table~\ref{tab:emotion-results},
\textbf{D} denotes the Dictator allocator, \textbf{UP} the Ultimatum
proposer, and \textbf{UR} the Ultimatum responder; \cmark, \xmark, and
\warnmark\ mark agreement, contradiction, and an ambiguous shift relative to
the human ground-truth direction. The \textbf{Simulation Score} awards two
points per \cmark, one per \warnmark, and zero per \xmark\ across the 15
emotion--role conditions (maximum 30). All game runs use temperature $0.7$
with 200 runs per condition.

\paragraph{Results.}
\textit{MetaSteer} outperforms the agentic reference on five of the six
settings: it improves the Simulation Score relative to \textsc{EAI-CP}
(Table~\ref{tab:emotion-results}) and the macro- and micro-averaged GAE
relative to \textsc{Seval-En} (Table~\ref{tab:gae-results}). These results
suggest that gains in steered text-generation performance are associated
with more controllable agents more broadly.

\begin{table*}[t]
    \caption{
    SocialEval GAE results across proself, antisocial, and prosocial targets, three model families, and two parameter scales each ($N=1210$ scenarios). Human baseline from 20 native Chinese/English graduate-student annotators (14 scenarios per language). Best score per metric per model in \textbf{bold}.
    }
  \centering
  \setlength{\tabcolsep}{3pt}
  \begin{tabular}{ll c cc cccc cc}
    \toprule
    \multirow{2}{*}{\textbf{Agent}} & \multirow{2}{*}{\textbf{Meth.}}
      & \textbf{Proself} & \multicolumn{2}{c}{\textbf{Antisocial}} & \multicolumn{4}{c}{\textbf{Prosocial}} & \multicolumn{2}{c}{\textbf{Overall GAE}} \\
    \cmidrule(lr){3-3} \cmidrule(lr){4-5} \cmidrule(lr){6-9} \cmidrule(lr){10-11}
    & & \textbf{C} & \textbf{I} & \textbf{C}
      & \textbf{C} & \textbf{A} & \textbf{N} & \textbf{A}
      & \textbf{Macro} & \textbf{Micro} \\
    \midrule

    \multicolumn{2}{l}{Human (avg)} & 40.00 & 60.00 & 40.00 & 60.00 & 55.00 & 55.00 & 70.00 & 55.16 & -- \\
    \midrule

    \multirow{2}{*}{\LlamaThreeTwoThreeB}
      & \textsc{Seval-En}       & \textbf{32.67} & 75.00 & 30.99 & \textbf{31.36} & \textbf{43.33} & 36.20 & 31.25 & 40.11 & 37.71 \\
      & \textit{MetaSteer}  & 26.49 & \textbf{85.71} & \textbf{50.00} & 23.44 & 35.29 & \textbf{49.06} & \textbf{36.24} & \textbf{43.75} & \textbf{39.10} \\
    \midrule
    \multirow{2}{*}{\LlamaThreeOneEightB}
      & \textsc{Seval-En}       & 28.57 & \textbf{87.14} & 50.00 & 22.73 & 35.50 & 41.28 & \textbf{52.35} & 45.37 & 39.93 \\
      & \textit{MetaSteer}  & \textbf{35.25} & 76.25 & \textbf{62.50} & \textbf{32.00} & \textbf{64.33} & \textbf{47.14} & 32.14 & \textbf{49.94} & \textbf{46.86} \\
    \midrule
    \multirow{2}{*}{\GemmaTwoTwoB}
      & \textsc{Seval-En}       & 25.16 & 50.00 & 26.25 & 25.00 & 35.00 & 49.75 & \textbf{25.00} & 33.74 & 33.43 \\
      & \textit{MetaSteer}  & \textbf{29.41} & \textbf{62.50} & \textbf{42.86} & \textbf{29.35} & \textbf{54.55} & \textbf{50.68} & 6.25  & \textbf{39.37} & \textbf{38.23} \\
    \midrule
    \multirow{2}{*}{\GemmaTwoNineB}
      & \textsc{Seval-En}       & \textbf{27.86} & 75.00 & \textbf{75.00} & 25.00 & \textbf{50.00} & 45.45 & 18.99 & 45.33 & 40.72 \\
      & \textit{MetaSteer}  & 26.67 & \textbf{100.00} & 62.50 & \textbf{37.80} & 41.67 & \textbf{50.00} & \textbf{36.67} & \textbf{50.76} & \textbf{45.49} \\
    \midrule
    \multirow{2}{*}{\QwenThreeOnePointSevenB}
      & \textsc{Seval-En}       & \textbf{43.40} & 50.00 & 26.25 & 23.20 & 38.89 & 41.10 & \textbf{32.67} & \textbf{36.50} & \textbf{36.32} \\
      & \textit{MetaSteer}  & 32.05 & \textbf{56.96} & \textbf{35.82} & \textbf{23.38} & \textbf{41.58} & \textbf{45.66} & 19.59 & 36.44 & 35.28 \\
    \midrule
    \multirow{2}{*}{\QwenThreeFourB} & \textsc{Seval-En} & 25.17 & 45.33 & 53.95 & 31.61 & 36.07 & \textbf{54.39} & 23.68 & 38.60 & 37.81 \\ & \textit{MetaSteer} & \textbf{31.97} & \textbf{56.41} & \textbf{58.75} & \textbf{34.72} & \textbf{37.10} & 42.17 & \textbf{35.85} & \textbf{42.43} & \textbf{39.78} \\

    \bottomrule
  \end{tabular}%
  \label{tab:gae-results}
\end{table*}

\section{Geometric Analysis}
\label{sec:geometry}

Prior work suggests that hidden-state trajectories encode information through
both absolute position and local geometric dynamics, including velocity and
curvature~\citep{xu2024geometry,park2025iclr,gjolbye2026reasoning,
zhou2026geometry}. We therefore test whether concept steering changes the
semantic location of a response while preserving aspects of its
instruction-associated trajectory dynamics.

We analyze 29,484 segments from 2,550 steered responses spanning 381
instructions, 70 concepts, and three genres (code, text, math). Segments are
encoded either cumulatively with their preceding context or independently.
Data construction, filtering, pooling, and similarity metrics are detailed in
Appendix~\ref{app:cross_model_similarity}.

\paragraph{Higher-order geometry is associated with instruction structure.}
Position similarity remains high across instruction, concept, and genre
groupings and changes little after shuffling, making it weakly diagnostic of
ordered structure. In contrast, instruction-grouped velocity increases from
$.06/.06$ to $.31/.38$ under cumulative encoding and from $.00/.00$ to
$.29/.37$ under isolated encoding. Acceleration similarly increases from
$.02/.02$ to $.27/.35$ and from $.00/.00$ to $.27/.35$, respectively.
Isolated Menger-curvature similarity increases from $.02/.00$ to $.29/.37$,
while the cumulative increase is weaker ($.32/.35$ to $.36/.44$) because
accumulated context retains shared structure after shuffling. Concept- and
genre-grouped similarities remain lower, indicating that velocity,
acceleration, and curvature partially reflect shared instruction structure.
Complete results appear in Appendix~\ref{app:cross_model_similarity},
Tables~\ref{tab:cumulative_similarity} and \ref{tab:isolated_similarity}, and
Figure~\ref{fig:three_trajectory_plots}\textbf{\captiona}--\textbf{\captionb}.

\paragraph{The geometric effect depends on the concept.}
In a complementary experiment, we compare the positional displacements
induced by \textit{MetaSteer} with those produced by CAA for 150 concepts proposed by~\citet{fan2026your}. Agreement is
quantified with the cosine alignment score $s_c$ (Eq.~\ref{eq:caa-alignment}),
averaged across six models; complete experimental details are provided in
Appendix~\ref{app:concept_linearity}. Figure~\ref{fig:three_trajectory_plots}\textbf{\captionc}
reports the mean CAA-alignment score for each concept, with shadow denoting one
standard deviation across models, and organizes the concepts into semantic
clusters. Concepts that induce broad \emph{global} changes, such as shifts in
language or output format, are more accurately represented by a linear
steering direction; concepts requiring context-dependent combinations of
local and global changes, such as discourse development or argument framing,
are less adequately captured by a single CAA direction.

\begin{figure*}[t]
    \centering

    \begin{minipage}[t]{0.3\textwidth}
        \centering
        \includegraphics[width=\linewidth]{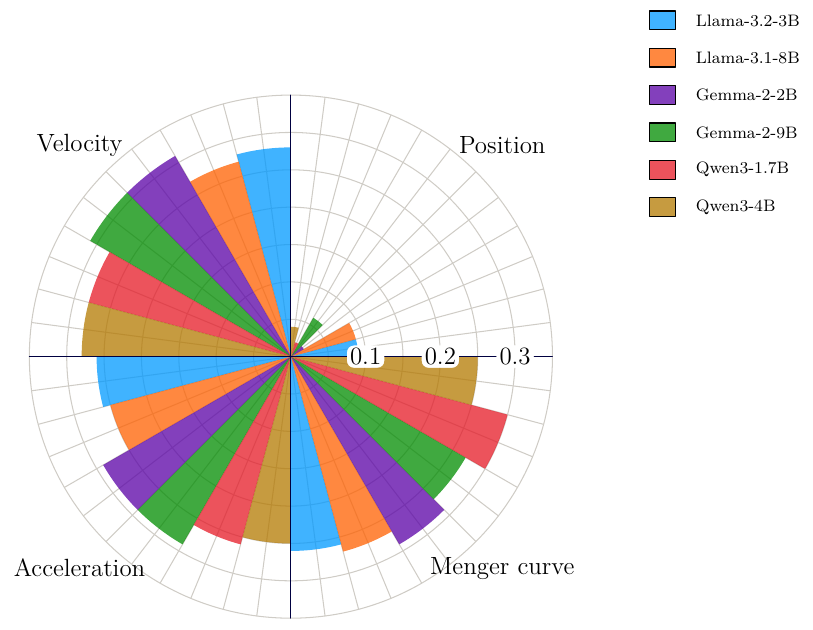}
        \captiona
    \end{minipage}
    \hfill
    \begin{minipage}[t]{0.3\textwidth}
        \centering
        \includegraphics[width=\linewidth]{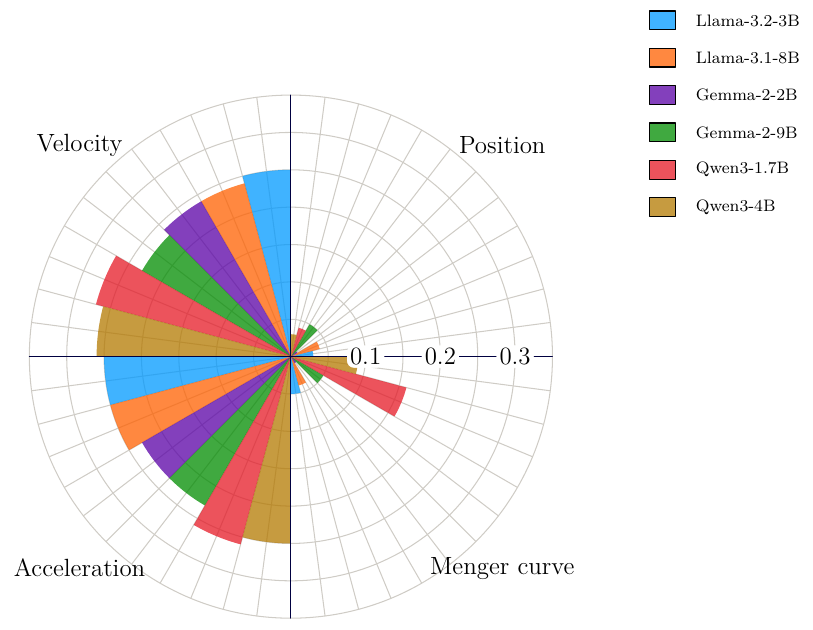}
        \captionb
    \end{minipage}
    \hfill
    \begin{minipage}[t]{0.34\textwidth}
        \centering
        \includegraphics[width=\linewidth]{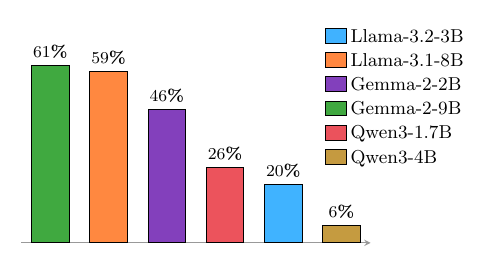}
        \captionc
    \end{minipage}

    \caption{\textbf{Trajectory geometry and safety.}
\textbf{\captiona} and \textbf{\captionb} show instruction-grouped gains in
trajectory similarity relative to a randomly shuffled control for position,
velocity, acceleration, and Menger curvature under isolated-segment \textbf{\captiona} and
cumulative-context \textbf{\captionb} encoding, respectively. \textbf{\captionc} shows obedience
rates for harmful steering concepts from JailbreakBench; higher values indicate
weaker refusal behaviour.}
    
\label{fig:radars}
\end{figure*}

\section{Supporting Analyses}
\label{sec:side-checks}

We conduct supporting analyses of safety, evaluator reliability, and capability retention. Because activation steering can weaken safety-relevant behavior such as refusal \citep{marks2023geometry,geiger2024finding,bao2026faithful,wu2026improved}, we assess whether MetaSteer can bypass refusals using harmful behaviours from JailbreakBench \citep{chao2024jailbreakbench}. Each behaviour is decomposed into a benign instruction and a risky steering concept, and generated responses are classified according to whether they comply with or refuse the harmful request. The results indicate that transferable steering can weaken refusal behaviour in some contexts (Figure~\ref{fig:radars}\textbf{\captionc}), motivating caution before deployment. Further details are provided in Appendix~\ref{app:safety}.

We also assess the reliability of LLM-based evaluation, which offers an efficient alternative to human assessment while raising concerns about consistency, calibration, and bias \citep{zheng2023judging,zhang2023wider,gu2024survey,yu2025ais,GU2026101253}. On selected samples from the text-generation evaluations, agreement among the LLM judges was strong. Detailed per-method, pairwise, and calibration analyses are reported in Appendix~\ref{app:extended_inter_rater}. Human-anchored validation of the judge scores is provided in Appendix~\ref{app:qualitative_examples}.
Finally, we evaluate capability retention by comparing the adapted models with their unmodified bases on MMLU \citep{hendrycks2020measuring} and TruthfulQA \citep{lin2022truthfulqa}. The results show generally limited changes in knowledge and truthfulness, with no consistent degradation across models. Complete per-model results are reported in Appendix~\ref{sec:capability_retention}.

\section{Conclusion}

We introduced \textsc{MetaSteer}, a preference-trained low-rank intervention
applied to all four attention projection matrices. Unlike a fixed activation
vector, its parameters remain fixed after training while its induced activation
effect varies with the input context. The method therefore addresses three
linked design questions: where to intervene, how to learn the intervention, and
how to avoid imposing a globally linear representation of each concept.

A single intervention trained on pooled preference data transfers to unseen
concepts and instruction combinations across three text-generation benchmarks
and three agentic settings without per-benchmark adaptation. Geometric
analyses further show that steering often changes trajectory position while partially preserving aspects of its local dynamics, including velocity and acceleration, although the strength of this effect varies across concepts.

Overall, our results support viewing steering as controlled trajectory
displacement: the intervention must induce a target behavior while preserving
task-relevant dynamics and general model utility. We additionally assess
evaluator reliability and identify safety considerations that should be
addressed before deploying transferable steering methods.





\subsection*{Reproducibility Statement}

The source code and pretrained model checkpoints will be made publicly available in an upcoming arXiv version shortly.

\subsubsection*{Acknowledgments}

This project is supported by the ARC Centre of Excellence for Automated Decision-Making and Society (CE200100005). Computational facilities were provided by the School of Computer Science and Engineering at UNSW Sydney through Katana.

\bibliography{iclr2027_conference.bib}
\bibliographystyle{iclr2027_conference}

\appendix

\section{Extended Related Work}
\label{app:extended_related_work}

Model steering can be organized as a design space defined by three related questions: what geometric structure is assumed for the target concept in activation space, where the intervention is applied, and how the intervention is computed or learned. This organization offers a more direct account of the literature than a division based solely on whether a method modifies the prompt, the activations, or the parameters: prompt- and parameter-based methods correspond primarily to different intervention locations, whereas activation-based methods additionally require an explicit choice of geometric assumption and learning procedure.

\subsection{Geometric Assumption}
\label{app:geometric_assumption}

\paragraph{Linear and feature-based representations.}
A dominant assumption in representation-level steering is that semantically meaningful concepts correspond approximately to directions, hyperplanes, or low-dimensional subspaces in activation space \citep{xiong2024everything,bereska2024mechanistic,saglam2025large}. Under this view, a desired behavior can be represented by a single vector added to the model's activation at a chosen layer and intensity. Contrastive Activation Addition (CAA) \citep{panickssery2023steering} is a prominent instance, constructing a steering vector from the difference between activations associated with desired and undesired behavior; related work shows that such directions can be composed and sometimes transfer across tasks, concepts, or models \citep{turner2023steering,karvonen2024emergent,nguyen2025multi}. Sparse-autoencoder (SAE) methods offer a related, feature-based instantiation of the linearity assumption: rather than manipulating undifferentiated coordinates of the residual stream, they identify sparse latent features intended to correspond to interpretable concepts \citep{lieberum2024gemma,he2024llama}, which have in turn been used to steer safety, fairness, truthfulness, instruction-following, and role-play behavior \citep{he2025towards,he2025saif,wang2025improving}. The reliability, disentanglement, and causal significance of these features nonetheless remain debated \citep{he2025sae,arad2025saes,cho2025corrsteer}.

\paragraph{Beyond the linear assumption.}
Recent work challenges the premise that concepts are represented by a single, fixed direction, showing instead that they may occupy curved, anisotropic, or otherwise non-Euclidean manifolds \citep{engels2025not,modell2025origins}. If a concept's representation varies with the input, a global vector cannot capture the intervention appropriate to every context, motivating nonlinear, context-conditioned steering procedures \citep{nguyen2026beyond,zhao2026odesteer,mishra2026steered} as well as more expressive transformations such as learned steering operators and rotations \citep{soo2025interpretable,vu2026angular}. The geometric question is therefore not only whether a concept admits a direction, but whether the appropriate transformation should vary across inputs, layers, or behavioral objectives. \emph{MetaSteer} learns a nonlinear, context-dependent intervention rather
than fixing a single global steering direction. We characterize the resulting
geometric variation empirically through displacement and trajectory analyses,
without assuming that the intervention's effective dimensionality equals the
intrinsic dimensionality of the hidden-state trajectory.

\subsection{Intervention Site}
\label{app:intervention_site}

The location of an intervention determines which part of the computation is modified and how directly it interacts with the model's existing representations. Steering methods range from interventions applied before generation begins to interventions applied within individual computational components or directly to model parameters.

\paragraph{Input- and parameter-level steering.}
At the input level, prompt-based steering modifies the conditions under which the model generates its answer: Chain-of-Thought prompting decomposes a problem into intermediate reasoning steps \citep{wei2022chain}, role-playing and persona-based prompts assign the model an identity or behavioral frame \citep{kong-etal-2024-better,miehling2025evaluating}, and AxBench shows that prompts generated by a stronger model can steer a weaker one \citep{wu2025axbench}. These approaches are easy to deploy because they leave parameters and activations untouched, but their effectiveness can depend on prompt-engineering expertise or on access to a stronger supervising model. At the parameter level, fine-tuning changes the model weights or introduces trainable modules that persist across inputs, including adapters \citep{pmlr-v97-houlsby19a}, prompt tuning \citep{lester2021power}, prefix tuning \citep{li2021prefix}, Low-Rank Adaptation (LoRA) \citep{hu2022lora}, and compact representation-level interventions such as Representation Fine-Tuning (ReFT) \citep{trung-etal-2024-reft}, with scaling-aware and stacked adaptation strategies proposed to reduce forgetting under continued adaptation \citep{kalajdzievski2024scaling,patil-etal-2025-stacked}. These methods make the intervention persistent and trainable, at the cost of additional training requirements and, potentially, reduced preservation of unrelated capabilities.

\paragraph{Internal activation sites.}
For runtime activation steering, the residual stream is the most common intervention site, as it is accessible throughout the forward pass and provides a shared space through which information propagates \citep{nguyen2025multi,hsu2026contextual}; this makes it convenient for vector addition, activation patching, and contextual regulation of steering strength, but modifying it can affect many downstream computations at once and so limit specificity. Recent work instead targets more localized components: MLP activations have been used to control particular behaviors or suppress spurious associations \citep{yu2026wasd,yan2026spurious}, and individual attention heads have been identified or manipulated to control behaviors such as sycophancy \citep{genadi2026sycophancy,luo2026don}, complementing activation-patching and causal-tracing tools that more generally locate components whose activations causally influence a target behavior \citep{zhang2024towards}. Attention projection matrices offer a further, more structured site: rather than adding a fixed vector after a layer has computed its representation, modifying a projection matrix changes how the current hidden state is transformed, so that the effect of the intervention depends on the representation being processed. The closest work to ours is \citet{luo2026don}, which intervenes only on the query projection; \emph{MetaSteer} instead intervenes jointly on all four attention projection matrices---query, key, value, and output---obtaining context-dependent behavior by construction rather than through an additive vector.

\paragraph{Feature-level locations.}
SAE-based steering can also be viewed as selecting a feature-level location within the representation space: dictionary-learning methods identify sparse features and modify the activation of a selected one \citep{lieberum2024gemma,he2024llama,wu2025axbench}, which can improve interpretability and offer more targeted control than modifying the full residual stream. However, SAE checkpoints are available for only a subset of model families, and the meaning and causal role of individual features can vary across models and tasks \citep{he2025sae,arad2025saes,cho2025corrsteer}; more broadly, the effectiveness of activation steering is known to be sensitive to the choice of model, task, layer, and component \citep{tan2024analysing}. These limitations motivate methods that exploit a structured intervention site without requiring a manually identified feature for every target concept.

\subsection{Learning Procedure}
\label{app:learning_procedure}

Once a site has been selected, a second problem is how to determine the transformation applied there. Existing methods differ in whether the intervention is specified heuristically, estimated from contrastive activations, learned as a causal operator, or optimized from behavioral or preference feedback.

\paragraph{Heuristic and contrastive construction.}
The simplest procedures use a manually chosen direction and tune only its magnitude at inference time. Contrastive methods such as CAA estimate a direction from paired examples of desired and undesired behavior \citep{panickssery2023steering}, while other approaches rely on heuristic transformations or statistical structure in activation space, including the PCA- and LAT-based methods surveyed in AxBench \citep{wu2025axbench}. These procedures are computationally inexpensive and effective when the target behavior is well described by a stable direction, but they require the practitioner to choose an appropriate layer, direction, and strength, and may not adapt well when the relevant representation shifts across queries.

\paragraph{Learned operators and contextual control.}
To increase flexibility, several methods learn the intervention itself rather than specifying a fixed vector: some learn to predict the causal effect of an intervention \citep{arad2025saes,cho2025corrsteer}, others replace vector addition with a learned steering operator \citep{soo2025interpretable,vu2026angular}, and contextual methods regulate the direction or intensity of steering according to the current input \citep{hsu2026contextual}. \citet{sun2025hypersteer} learns steering vectors end-to-end via cross-attention over semantic embeddings, but extracts them using a separate, architecturally identical model; \emph{MetaSteer} instead uses the same model throughout. These approaches improve expressiveness relative to heuristic construction, though their performance can depend on task-specific training data, high-quality representations, or carefully designed auxiliary modules.

\paragraph{Objective-based and preference-based learning.}
A complementary line of work formulates steering as an optimization problem that balances the desired behavioral change against the preservation of general model utility, tuning both intervention orientation and intensity to trade off steering effectiveness $\mathbb{S}$ against utility $\mathbb{U}$ \citep{aravindan2026opium,nguyen2026minimizing,luo2026learning}, making explicit that a stronger intervention is not necessarily preferable if it damages unrelated capabilities. Preference-based methods provide a related but distinct learning signal: rather than manually specifying an activation direction, they use comparisons between preferred and non-preferred behaviors to learn how the model should be steered, for instance by cloning desired behavior directly in activation space \citep{raina2025d}. Such objectives make the learning signal more directly behavioral, but remain constrained by the representation through which the preference signal is transmitted; a fixed representation can in particular create an information bottleneck when it cannot express every behavioral distinction the target task requires \citep{raina2025d}. \emph{MetaSteer} is closest in learning approach to \citet{raina2025d}, building on preference optimization to learn a transferable policy over interventions, but avoids this bottleneck by learning nonlinear, context-dependent interventions rather than a single low-dimensional vector per concept.

\paragraph{Efficiency and transferability.}
These procedures impose different data and computational requirements: heuristic and contrastive methods are inexpensive but may require manual tuning or concept-specific examples; learned operators and parameter-efficient adaptations are more flexible but require additional training and may specialize to the tasks seen during training; and preference-based methods offer a more direct behavioral objective whose generalization depends on the coverage and quality of the preference data. These trade-offs motivate methods that learn expressive interventions from limited supervision while still transferring to unseen concepts and tasks.

\subsection{Evaluation}
\label{app:evaluation}

Although evaluation is not itself an intervention dimension, it determines how improvements along the geometric, site, and learning dimensions are measured. The increasing capability of frontier language models, together with the cost of human annotation, has motivated using LLMs as proxies for human evaluators \citep{zheng2023judging}; such LLM-as-a-Judge (LaaJ) frameworks are now widely used in automated evaluation pipelines, including in multi-agent settings \citep{zhang2023wider,gu2024survey,GU2026101253,yu2025ais}, and prior work reports substantial agreement between LLM judges and human annotators while flagging open concerns around reliability, fairness, reproducibility, calibration, and systematic bias \citep{wang-etal-2024-large-language-models-fair,gu2024survey,GU2026101253,yu2025ais}. These concerns bear directly on steering research, where evaluation typically relies on a single frontier model as judge and, in some cases, the same model family that contributed training or evaluation data also serves as evaluator, creating the possibility of systematic bias toward behaviors that family favors \citep{wu2025axbench}; prior studies of activation steering and SAE-based interventions have adopted this paradigm \citep{wu2025axbench,soo2025interpretable,arad2025saes,turner2023steering}, despite the reliability and robustness of LLM judges for measuring steering quality remaining insufficiently characterized. This motivates the careful assessment of automated evaluation alongside the steering results we report.

\subsection{Extension to Multi-Head, Grouped-Query Attention, and RoPE}
\label{app:proofs-multihead}

The derivation above treats a single attention head with no positional
encoding. We now give the general form used by every backbone in this
work, each of which combines grouped-query attention with RoPE.

\paragraph{Setup.} Let $H$ denote the number of query heads and
$H_{kv}\leq H$ the number of key/value heads, with group size
$g=H/H_{kv}$; query head $h$ shares key/value head
$\kappa(h)=\lceil h/g\rceil$. Standard multi-head attention is the special
case $H_{kv}=H$. Per-head baseline projections are
\[
\mathbf{q}_i^{(h)}=\mathbf{z}_i\mathbf{W}_q^{(h)},\qquad
\mathbf{k}_j^{(\kappa)}=\mathbf{z}_j\mathbf{W}_k^{(\kappa)},\qquad
\mathbf{v}_j^{(\kappa)}=\mathbf{z}_j\mathbf{W}_v^{(\kappa)},
\]
the corresponding column-blocks of $\mathbf{W}_q,\mathbf{W}_k,\mathbf{W}_v$,
with adapters $\Delta\mathbf{W}_q^{(h)},\Delta\mathbf{W}_k^{(\kappa)}$ the
matching blocks of $\Delta\mathbf{W}_q,\Delta\mathbf{W}_k$ from
Eq.~\ref{eq:lora-adapters}, since LoRA is applied to the full projection
matrix prior to reshaping into heads.

\paragraph{RoPE rotation.} Before the dot product, RoPE applies a
position-dependent orthogonal rotation $\mathbf{R}_{\Theta,m}\in
\mathbb{R}^{d_h\times d_h}$ to each head's query and key vectors,
$\hat{\mathbf{q}}_i^{(h)}=\mathbf{q}_i^{(h)}\mathbf{R}_{\Theta,i}$ and
$\hat{\mathbf{k}}_j^{(\kappa)}=\mathbf{k}_j^{(\kappa)}\mathbf{R}_{\Theta,j}$,
satisfying the defining relative-position identity
$\mathbf{R}_{\Theta,i}\mathbf{R}_{\Theta,j}^\top=\mathbf{R}_{\Theta,i-j}$.
The per-head logit is therefore
\[
s_{ij}^{(h)}
=
\frac{1}{\sqrt{d_h}}\,\mathbf{q}_i^{(h)}\mathbf{R}_{\Theta,i-j}
\big(\mathbf{k}_j^{(\kappa(h))}\big)^\top,
\]
depending only on the relative offset $i-j$.

\paragraph{Per-head perturbation.} Repeating the expansion of
Equations~\ref{eq:term1}--\ref{eq:term3} per head, with
$\mathbf{R}_{\Theta,i-j}$ inserted between the query-side and key-side
factor exactly as above, the three cross terms again collapse into a
single bilinear form, now indexed by head and relative position:
\begin{equation}
\delta_{ij}^{(h)}
=
\frac{1}{\sqrt{d_h}}\,\mathbf{z}_i\,\mathbf{M}^{(h)}(i-j)\,\mathbf{z}_j^\top,
\label{eq:M-def-multihead}
\end{equation}
\begin{equation}
\begin{aligned}
\mathbf{M}^{(h)}(m) :=\ &
\lambda_k\,\mathbf{W}_q^{(h)}\mathbf{R}_{\Theta,m}\mathbf{B}_k^{(\kappa(h))}\mathbf{A}_k^{(\kappa(h))\top}\\
&+\lambda_q\,\mathbf{A}_q^{(h)}\mathbf{B}_q^{(h)\top}\mathbf{R}_{\Theta,m}\mathbf{W}_k^{(\kappa(h))\top}\\
&+\lambda_q\lambda_k\,\mathbf{A}_q^{(h)}\mathbf{B}_q^{(h)\top}\mathbf{R}_{\Theta,m}\mathbf{B}_k^{(\kappa(h))}\mathbf{A}_k^{(\kappa(h))\top}.
\end{aligned}
\label{eq:M-def-rope}
\end{equation}
Equation~\ref{eq:M-def} is recovered exactly when $H=H_{kv}=1$ and
$\mathbf{R}_{\Theta,m}=\mathbf{I}$. Since $\mathbf{R}_{\Theta,m}$ is a fixed
orthogonal matrix independent of $\mathbf{z}_i,\mathbf{z}_j$, the
genericity argument following Equation~\ref{eq:grad-full} applies
unchanged to each $\mathbf{M}^{(h)}(m)$: both partial derivatives are
non-zero except on a measure-zero set, now for every head $h$ and every
relative offset $m$ realized by the context window.

\paragraph{GQA-induced coupling.} Under grouped-query attention
($H_{kv}<H$), the $g$ query heads sharing key/value head $\kappa$ have
distinct $\mathbf{M}^{(h)}(m)$ (through $\mathbf{W}_q^{(h)},\mathbf{A}_q^{(h)},\mathbf{B}_q^{(h)}$)
but share the same key-side adapter $\mathbf{A}_k^{(\kappa)},\mathbf{B}_k^{(\kappa)}$:
a single trained key-side update therefore perturbs $g$ heads'
attention patterns simultaneously, each through a different query-side
projection.

\section{Derivation of the Logit-Gap Identity}
\label{app:logit-gap}

Equation~\ref{eq:softmax-def} states that, under a softmax output parameterization, the log-probability gap between two completions is linear in the final hidden state. This identity is derived by \citet{raina2025d} (Section~2.2) and underlies the single-vector characterization of DPO recalled in Section~\ref{sec:background}. We reproduce the derivation here for completeness.

\paragraph{Setup.} Under standard softmax parameterization, the model's conditional distribution over the next token $y$ is
\begin{equation}
\begin{aligned}
\pi(y \mid x) &= \frac{\exp\big(z_y(x)\big)}{\sum_{y'} \exp\big(z_{y'}(x)\big)}, \\
z_y(x) &= \langle \mathbf{h}(x), \mathbf{e}_y \rangle,
\end{aligned}
\label{eq:softmax-def}
\end{equation}
where $\mathbf{h}(x) \in \mathbb{R}^d$ is the model's final hidden state for prompt $x$, and $\mathbf{e}_y$ is the output (unembedding) vector for token $y$, so that the logit $z_y(x)$ is their inner product.

\paragraph{Log-probability of a single completion.} Taking the logarithm of Equation~\ref{eq:softmax-def},
\begin{equation}
\begin{aligned}
\log \pi(y \mid x) &= z_y(x) - \log \sum_{y'} \exp\big(z_{y'}(x)\big) \\
&= \langle \mathbf{h}(x), \mathbf{e}_y\rangle - \log Z(x),
\end{aligned}
\label{eq:log-softmax}
\end{equation}
where $Z(x) = \sum_{y'} \exp\langle \mathbf{h}(x), \mathbf{e}_{y'}\rangle$ is the partition function. Crucially, $Z(x)$ depends only on $x$ (through $\mathbf{h}(x)$) and on the full vocabulary, \emph{not} on the particular token $y$ whose probability is being evaluated.

\paragraph{Cancellation of the partition function.} Consider two completions, $y^+$ and $y^-$, evaluated at the same prompt $x$ (and hence the same hidden state $\mathbf{h}(x)$ and the same $Z(x)$). Applying Equation~\ref{eq:log-softmax} to each and subtracting,
\begin{equation}
\begin{gathered}
\log \pi(y^+ \mid x) - \log \pi(y^- \mid x) \\
= \Big[\langle \mathbf{h}(x), \mathbf{e}_{y^+}\rangle - \log Z(x)\Big] - \Big[\langle \mathbf{h}(x), \mathbf{e}_{y^-}\rangle - \log Z(x)\Big] \\
= \langle \mathbf{h}(x), \mathbf{e}_{y^+}\rangle - \langle \mathbf{h}(x), \mathbf{e}_{y^-}\rangle,
\end{gathered}
\label{eq:cancel}
\end{equation}
since $\log Z(x)$ is identical in both terms and cancels exactly. This cancellation is the entire content of the result: it holds regardless of vocabulary size, temperature, or the specific values of $\mathbf{h}(x)$, because it follows purely from the algebraic structure of the softmax, not from any property of the trained model.

\paragraph{Linearity in the hidden state.} By bilinearity of the inner product, Equation~\ref{eq:cancel} simplifies to
\begin{equation}
\begin{gathered}
\log \pi(y^+ \mid x) - \log \pi(y^- \mid x) \\
= \big\langle \mathbf{h}(x),\ \mathbf{e}_{y^+} - \mathbf{e}_{y^-} \big\rangle \\
= \langle \mathbf{h}(x), \mathbf{v}\rangle, \qquad \mathbf{v} := \mathbf{e}_{y^+} - \mathbf{e}_{y^-},
\end{gathered}
\label{eq:logit-gap-derived}
\end{equation}
which is exactly Equation~\ref{eq:logit-gap}. The gap is therefore an inner product between the hidden state and a single, context-independent vector $\mathbf{v}$ determined only by the output embeddings of the two completions being compared.

\paragraph{Consequence for the DPO gradient.} Since the DPO loss is a function of $\rho_\theta(y^+\mid x) - \rho_\theta(y^-\mid x)$, and by Equation~\ref{eq:logit-gap-derived} this quantity is linear in $\mathbf{h}(x)$ with fixed slope $\mathbf{v}$, its gradient with respect to the hidden state is
\begin{equation}
\nabla_{\mathbf{h}(x)} \mathcal{L}_{\text{DPO}} = -\beta\,\sigma\!\big(-\beta\langle \mathbf{h}(x), \mathbf{v}\rangle + \text{const}\big)\, \mathbf{v} \ \propto\ -\mathbf{v},
\label{eq:grad-derived}
\end{equation}
where the scalar prefactor depends on $x$ but the direction does not: for every prompt $x$, the gradient points along the same vector $\mathbf{v}$, up to sign and magnitude. This is the source of the rank-one/single-direction characterization of single-concept DPO steering discussed in Section~\ref{sec:background} \citep{raina2025d}.

\section{Full Expansion of the Query-Key Perturbation}
\label{app:proofs}

This appendix expands Equation~\ref{eq:delta-ij} and derives the partial derivatives in Equation~\ref{eq:nonzero-grad} in full, showing that the query-key perturbation $\delta_{ij}$ reduces to a single bilinear form in $\mathbf{z}_i$ and $\mathbf{z}_j$, and that its dependence on each is generically non-zero.

\paragraph{Setup.} Recall $\mathbf{q}_i = \mathbf{z}_i\mathbf{W}_q$, $\mathbf{k}_j = \mathbf{z}_j\mathbf{W}_k$, $\Delta\mathbf{q}_i = \lambda_q\mathbf{z}_i\mathbf{A}_q\mathbf{B}_q^\top$, $\Delta\mathbf{k}_j = \lambda_k\mathbf{z}_j\mathbf{A}_k\mathbf{B}_k^\top$, with $\mathbf{z}_i,\mathbf{z}_j$ treated as row vectors and $\mathbf{W}_q,\mathbf{W}_k,\mathbf{A}_f\mathbf{B}_f^\top \in \mathbb{R}^{d\times d}$ (we take $d=d'$ for notational simplicity; the argument is unchanged for $d\neq d'$). The three terms of Equation~\ref{eq:delta-ij} are, expanding each inner product as a matrix product with a transpose,

\begin{equation}
\begin{aligned}
\langle \mathbf{q}_i, \Delta\mathbf{k}_j\rangle
&= \lambda_k\, \mathbf{z}_i\mathbf{W}_q \big(\mathbf{z}_j\mathbf{A}_k\mathbf{B}_k^\top\big)^\top \\
&= \lambda_k\, \mathbf{z}_i \big(\mathbf{W}_q\mathbf{B}_k\mathbf{A}_k^\top\big) \mathbf{z}_j^\top
\end{aligned}
\label{eq:term1}
\end{equation}

\begin{equation}
\begin{aligned}
\langle \Delta\mathbf{q}_i, \mathbf{k}_j\rangle
&= \lambda_q\, \mathbf{z}_i\mathbf{A}_q\mathbf{B}_q^\top \big(\mathbf{z}_j\mathbf{W}_k\big)^\top \\
&= \lambda_q\, \mathbf{z}_i \big(\mathbf{A}_q\mathbf{B}_q^\top\mathbf{W}_k^\top\big) \mathbf{z}_j^\top
\end{aligned}
\label{eq:term2}
\end{equation}

\begin{equation}
\begin{aligned}
\langle \Delta\mathbf{q}_i, \Delta\mathbf{k}_j\rangle
&= \lambda_q\lambda_k\, \mathbf{z}_i\mathbf{A}_q\mathbf{B}_q^\top \big(\mathbf{z}_j\mathbf{A}_k\mathbf{B}_k^\top\big)^\top \\
&= \lambda_q\lambda_k\, \mathbf{z}_i \big(\mathbf{A}_q\mathbf{B}_q^\top\mathbf{B}_k\mathbf{A}_k^\top\big) \mathbf{z}_j^\top
\label{eq:term3}
\end{aligned}
\end{equation}
Each term is a scalar of the identical form $\mathbf{z}_i(\cdot)\mathbf{z}_j^\top$, differing only in the $d\times d$ matrix sandwiched between $\mathbf{z}_i$ and $\mathbf{z}_j^\top$.

\paragraph{Collapse to a single bilinear form.} Summing Equations~\ref{eq:term1}--\ref{eq:term3} per Equation~\ref{eq:delta-ij} and factoring out the common $\mathbf{z}_i(\cdot)\mathbf{z}_j^\top$ structure,

\begin{equation} 
\begin{gathered} \delta_{ij} = \frac{1}{\sqrt{d}}\, \mathbf{z}_i\, \mathbf{M}\, \mathbf{z}_j^\top, \\ \mathbf{M} := \lambda_k\mathbf{W}_q\mathbf{B}_k\mathbf{A}_k^\top + \lambda_q\mathbf{A}_q\mathbf{B}_q^\top\mathbf{W}_k^\top + \lambda_q\lambda_k\mathbf{A}_q\mathbf{B}_q^\top\mathbf{B}_k\mathbf{A}_k^\top, \end{gathered}
\label{eq:M-def} \end{equation}

where $\mathbf{M}\in\mathbb{R}^{d\times d}$ depends only on the frozen weights and the trained adapters $\{\mathbf{A}_q,\mathbf{B}_q,\mathbf{A}_k,\mathbf{B}_k\}$ and intensities $\lambda_q,\lambda_k$ --- it does \emph{not} depend on the token positions $i,j$. The entire query-key perturbation, despite being built from three separate inner products, is therefore exactly a single bilinear form in the two token activations $\mathbf{z}_i$ and $\mathbf{z}_j$, mediated by the fixed matrix $\mathbf{M}$.

\paragraph{Partial derivatives.} Equation~\ref{eq:M-def} is linear in each argument holding the other fixed, so its gradients follow directly from the bilinear-form identity $\partial(\mathbf{z}_i\mathbf{M}\mathbf{z}_j^\top)/\partial\mathbf{z}_i = \mathbf{M}\mathbf{z}_j^\top$ and $\partial(\mathbf{z}_i\mathbf{M}\mathbf{z}_j^\top)/\partial\mathbf{z}_j = \mathbf{M}^\top\mathbf{z}_i^\top$:
\begin{equation}
\frac{\partial \delta_{ij}}{\partial \mathbf{z}_i} = \frac{1}{\sqrt{d}}\,\mathbf{M}\,\mathbf{z}_j^\top, \qquad
\frac{\partial \delta_{ij}}{\partial \mathbf{z}_j} = \frac{1}{\sqrt{d}}\,\mathbf{M}^\top\mathbf{z}_i^\top.
\label{eq:grad-full}
\end{equation}

\paragraph{Genericity.} Equation~\ref{eq:grad-full} vanishes only if $\mathbf{M}\mathbf{z}_j^\top = \mathbf{0}$ (respectively $\mathbf{M}^\top\mathbf{z}_i^\top=\mathbf{0}$), which requires either (a) $\mathbf{M}=\mathbf{0}$ --- an exact, measure-zero cancellation of the three adapter terms in Equation~\ref{eq:M-def} that is not enforced by, and would not be a stable fixed point of, gradient-based training on a non-trivial preference-optimization objective --- or (b) $\mathbf{z}_j$ (resp.\ $\mathbf{z}_i$) lying in the null space of $\mathbf{M}$ (resp.\ $\mathbf{M}^\top$), a codimension-$\geq 1$ subset of activation space that generic hidden states do not occupy. Excluding these non-generic cases, both partial derivatives in Equation~\ref{eq:grad-full} are non-zero, establishing Equation~\ref{eq:nonzero-grad}.

\paragraph{Interpretation.} Equation~\ref{eq:grad-full} makes explicit \emph{why} the intervention is context-dependent rather than a fixed offset: the sensitivity of $\delta_{ij}$ to the query token $\mathbf{z}_i$ is not a constant vector, but $\mathbf{M}\mathbf{z}_j^\top$ --- a vector that itself changes with the key token $\mathbf{z}_j$ it is being compared against, and hence with whatever content (including the concept specification $c$) that key token encodes. This is the precise sense in which the effective steering direction is a function of the surrounding sequence rather than a parameter fixed at the end of training: the same query activation $\mathbf{z}_i$ receives a different perturbation gradient depending on which key $\mathbf{z}_j$ it attends to, mediated entirely through the single learned matrix $\mathbf{M}$.

\newtcolorbox{generatedresponse}{
    breakable,
    colback=gray!10,
    colframe=gray!45,
    boxrule=0.4pt,
    arc=2pt,
    left=6pt,
    right=6pt,
    top=5pt,
    bottom=5pt,
    before skip=6pt,
    after skip=10pt
}

\newcommand{\responseitem}[1]{%
    \noindent #1\par\vspace{0.35em}
}

\section{Extended Inter-Rater Reliability Analysis}
\label{app:extended_inter_rater}

This appendix extends the reliability analysis summarized in the main text.
A separate, human-anchored validation of judge scores is reported in
Appendix~\ref{app:quantitative_analysis}.

\paragraph{Setup.} We evaluate two tasks: the steering-evaluation protocol
of \citet{wu2025axbench} (steering-score and language-quality metrics) and
the open-ended generation protocol of \citet{soo2025interpretable}. Each is
scored independently by three judges -- \GPTFourOMini, \GeminiFlashLite, and
\ClaudeHaiku -- on steering success and output quality.

\paragraph{Overall and per-method agreement.}
Pooled across all method$\times$item units, judges agree strongly
(Krippendorff's $\alpha=0.81$; ICC(3,1)$=0.81$). Table~\ref{tab:inter-rater-reliability}
breaks this down by method on \GemmaTwoTwoB (layer 10): \textsc{LoRA} and
\textsc{LoReFT} show the strongest, most consistent agreement;
\textsc{PromptSteering}, \textsc{LAT}, and \textsc{DiffMean} are
intermediate; \textsc{PCA} and \textsc{DPO} are lowest\footnote{DPO outputs
may convey concepts more subtly, making binary scoring harder.}.

\begin{figure}[t]
  \includegraphics[width=\linewidth]{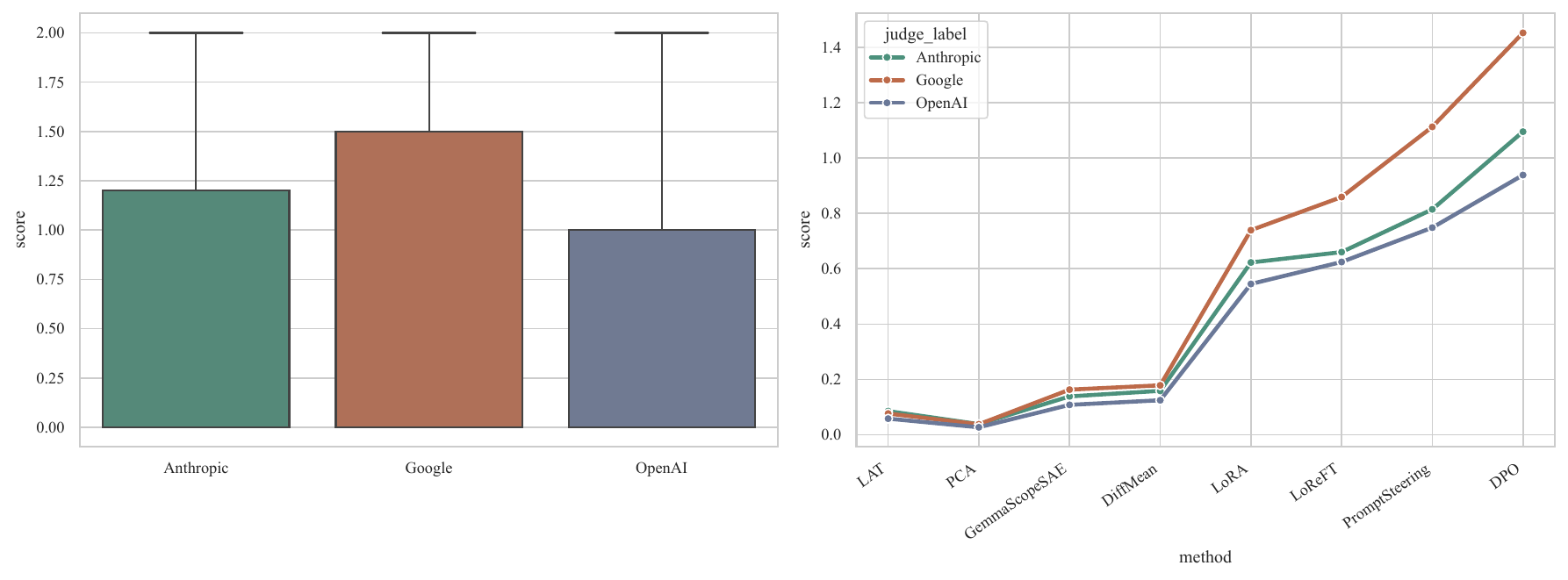}
  \caption{Judge score distributions, steering evaluation task.}
  \label{fig:t1_judge_bias}
\end{figure}

\begin{table}[t]
\caption{Inter-judge agreement across steering methods for the steering
evaluation task on \GemmaTwoTwoB (layer 10).}
\centering
\small
\begin{tabular}{lcc}
\hline
\textbf{Method} & \textbf{Krippendorff $\alpha$} & \textbf{ICC(3,1)} \\
\hline
\textsc{LoRA}          & 0.83 & 0.85 \\
\textsc{LoReFT}        & 0.77 & 0.81 \\
\textsc{LAT}           & 0.71 & 0.72 \\
\textsc{DiffMean}      & 0.71 & 0.71 \\
\textsc{PromptSteering}& 0.65 & 0.79 \\
\textsc{PCA}           & 0.54 & 0.56 \\
\textsc{DPO}           & 0.33 & 0.51 \\
\hline
\textbf{Pooled (all units)} & \textbf{0.81} & \textbf{0.81} \\
\hline
\end{tabular}
\label{tab:inter-rater-reliability}
\end{table}
\begin{figure}[t]
  \includegraphics[width=\linewidth]{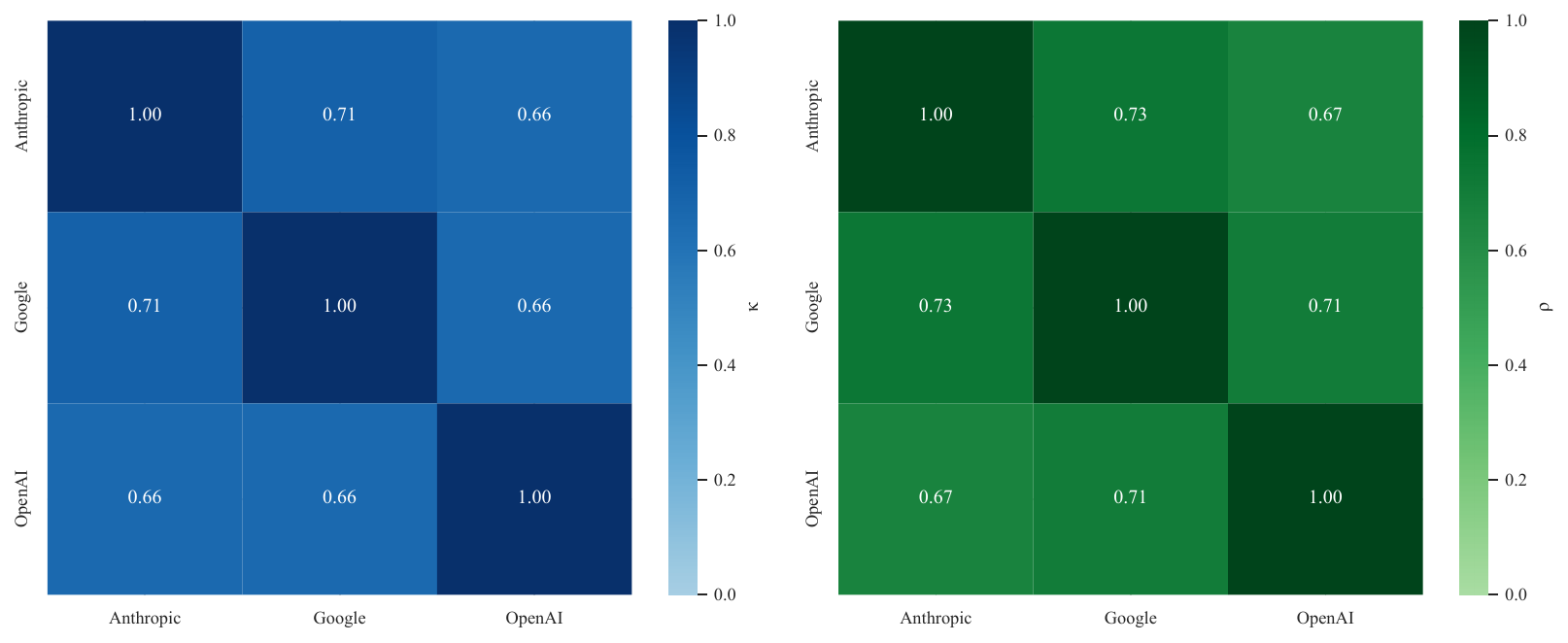}
  \caption{Macro-averaged pairwise judge agreement, steering evaluation task.}
  \label{fig:t1_pairwise_agreement}
\end{figure}
\paragraph{Pairwise agreement and calibration.}
Weighted Cohen's $\kappa$ and Spearman correlations
(Fig.~\ref{fig:t1_pairwise_agreement}) show moderate-to-good pairwise
agreement: judges are consistent in relative ordering but not fully
interchangeable at fine granularity, particularly for \textsc{DPO}. This
pattern holds on the open-ended task as well. Judges also differ
systematically in scale (Fig.~\ref{fig:t1_judge_bias}): one is consistently
more lenient (e.g., on \textsc{DPO} and \textsc{PromptSteering}), another
more conservative, confirmed by a mixed-effects model with significant fixed
effects. These differences are largely scale effects rather than
disagreement in ordering -- a variance decomposition on the open-ended task
attributes most variance to concept-level differences, with smaller
contributions from model and judge, and method rankings remain stable
across judges on both tasks.

\paragraph{Summary.}
Judges disagree modestly in absolute calibration but strongly agree in
relative ordering, supporting LLM-as-a-Judge as a reliable evaluation
strategy in our setting.

\section{Human Validation of LLM Judges}
\label{app:human_validation}

The increasing capability of frontier language models, together with the
high cost of human annotation, has motivated using LLMs as proxies for
human evaluators \citep{zheng2023judging,zhang2023wider,gu2024survey,GU2026101253,yu2025ais}.
While prior work reports substantial LLM-human agreement \citep{zheng2023judging},
concerns remain about reliability, fairness, reproducibility, and systematic
bias \citep{wang-etal-2024-large-language-models-fair,gu2024survey,GU2026101253,yu2025ais},
particularly in model-steering research, where a single frontier model
often both generates and evaluates outputs \citep{wu2025axbench,soo2025interpretable,arad2025saes,turner2023steering}.
We therefore cross-check our own evaluation setup against human judgment
below, and follow \citet{wu2023large} in assigning the LLM judge an explicit
evaluator role via prompting.

\subsection{Quantitative Analysis}
\label{app:quantitative_analysis}

We sampled 50 generated outputs (25 from \citet{wu2025axbench}, 25 from
\citet{soo2025interpretable}) and scored each, under an identical prompt,
with both \GPTFourOMini and a human annotator (a postgraduate student and
second-language English speaker). Agreement between the two sets of ratings
was moderately strong, with ICC(3,1) $\approx 0.70$, supporting
LLM-as-a-Judge as a reasonable proxy for human evaluation in our setting.
Representative side-by-side judgments are given in
Appendix~\ref{app:qualitative_examples}.

\subsection{Qualitative Examples}
\label{app:qualitative_examples}

We show a sample of human and LLM judgments for two identical inputs below.

\begin{tcolorbox}[title=\textbf{AxBench Sample}, colback=gray!5, colframe=black!60, breakable]
\textbf{Steering Concept:} Scientific terms related to research findings and implications \\
\textbf{Instruction:} What are different drawers I should have for clothes?

\vspace{4pt}
\textbf{Prompt Steering Output}
\begin{quote}
\small
While the question of optimal drawer organization isn't directly related to scientific research, let's view it through a lens of anthropological insights into human behavior and organizational psychology. Understanding why we organize our belongings is fascinating! One might draw on heuristic decision-making models. Our brains are naturally efficient and seek shortcuts to navigate everyday activities, and choosing a drawer organization system is an example of this. A structured system like this can make accessing and putting away clothes more efficient -- reducing cognitive load and increasing self-efficacy.
\end{quote}
\begin{itemize}
    \item Fluency: LLM = 1, Human = \texttt{2}
    \item Concept Relevance: LLM = 1, Human = \texttt{1}
    \item Instruction Relevance: LLM = 1, Human = \texttt{1}
\end{itemize}

\textbf{DPO Output}
\begin{quote}
\small
The scientific literature on organizational behaviors suggests that the strategic placement of clothing compartments, such as drawers, can significantly impact efficiency within an individual's wardrobe system. Recent studies indicate a correlation between drawer design and optimal apparel access, facilitating streamlined clothing selection based on contextual factors like occasion and activity. This finding emphasizes the necessity for tailored drawer configurations that align with personal preferences, offering a personalized approach to optimizing storage and retrieval, ultimately enhancing overall time management in the context of daily dressing routines. \ldots
\end{quote}
\begin{itemize}
    \item Fluency: LLM = 2, Human = \texttt{2}
    \item Concept Relevance: LLM = 2, Human = \texttt{2}
    \item Instruction Relevance: LLM = 2, Human = \texttt{1}
\end{itemize}
\end{tcolorbox}

\begin{tcolorbox}[title=\textbf{Free-Text Generation Sample (Base Model, \citet{soo2025interpretable})}, colback=gray!5, colframe=black!60, breakable]
\textbf{Steering Concept:} Anger \\
\textbf{Prompt:} I think \ldots

\vspace{4pt}
\textbf{Prompt Steering Output}
\begin{quote}
\small
\ldots I was just so mad at this restaurant for going overboard on the lack of customer service earlier this week, so after getting out of therapy I lost my, \ldots
\end{quote}
\begin{itemize}
    \item Steering Score: LLM = 0.67, Human = \texttt{0.75}
    \item Coherence: LLM = 0.67, Human = \texttt{0.60}
    \item Product: LLM = 0.45, Human = \texttt{0.45}
\end{itemize}

\textbf{DPO Output}
\begin{quote}
\small
\ldots wrath is the driving force behind my actions, fueled by resentment and frustration. Like a roaring inferno, it ignites my passion, shaping \ldots
\end{quote}
\begin{itemize}
    \item Steering Score: LLM = 0.78, Human = \texttt{0.85}
    \item Coherence: LLM = 0.78, Human = \texttt{0.70}
    \item Product: LLM = 0.61, Human = \texttt{0.60}
\end{itemize}
\end{tcolorbox}

These examples illustrate that LLM-as-a-Judge can serve as a reliable proxy
for human evaluation in the text-generation setting.

\section{Performance Degradation Analysis}
\label{sec:capability_retention}

One of the main limitations of steering a large language model toward a particular behaviour is an unwanted side effect on the model's general capabilities.
These side effects can range from degraded inherent safety guardrails~\cite{bao2026faithful} to drops of $2$--$4\%$ on general-capability benchmarks~\citep{ostermann2026weights,rimsky2024steering}.
We therefore run a side-check experiment to verify that MetaSteer does not induce significant performance degradation.

\paragraph{Testing scenario.}
For each backbone we compare two variants under an identical evaluation protocol: the original HuggingFace model (\emph{Base}) and the same model with the trained MetaSteer LoRA adapter applied (\emph{MetaSteer}).
Evaluations use the EleutherAI \texttt{lm-evaluation-harness} with multiple-choice log-likelihood scoring (not generative decoding).
We report accuracy on three held-out capability / truthfulness suites, using the full official test sets:
\begin{itemize}
    \item \textbf{MMLU}: broad knowledge and reasoning;
    \item \textbf{TruthfulQA MC1}: single-true-answer truthfulness;
    \item \textbf{TruthfulQA MC2}: multi-true-answer truthfulness.
\end{itemize}
Each task is evaluated at $k\in\{0,2,4\}$ few-shot examples.
Chat templates are disabled, and Qwen3 thinking mode is turned off, so scores reflect standard multiple-choice log-likelihood rather than chain-of-thought generation.
The reported delta is
\[
\Delta \;=\; \mathrm{Acc}(\text{MetaSteer}) - \mathrm{Acc}(\text{Base}),
\]
so negative values indicate degradation.

\paragraph{Results.}
Table~\ref{tab:capability_retention} summarises Base vs.\ MetaSteer accuracy (\%).
Across the evaluated models, MMLU changes are generally small. The largest
TruthfulQA drops occur for \QwenThreeZeroPointSixB, reaching 1.51 percentage points; among
the six primary backbones, the largest drop is 0.37 points, while
\GemmaTwoTwoB and \QwenThreeFourB improve on some TruthfulQA settings.

\begin{table*}[t]
\centering
\small
\setlength{\tabcolsep}{3.5pt}
\caption{Capability retention after MetaSteer.
Accuracy (\%) of the unsteered base model vs.\ the MetaSteer LoRA on MMLU and TruthfulQA (MC1/MC2) at $k=0,2,4$ shots.
$\Delta$ is MetaSteer $-$ Base in percentage points (negative $=$ degradation).}
\label{tab:capability_retention}
\begin{tabular}{ll ccc ccc ccc}
\toprule
& & \multicolumn{3}{c}{MMLU} & \multicolumn{3}{c}{TruthfulQA MC1} & \multicolumn{3}{c}{TruthfulQA MC2} \\
\cmidrule(lr){3-5}\cmidrule(lr){6-8}\cmidrule(lr){9-11}
Model & $k$ & Base & MS & $\Delta$ & Base & MS & $\Delta$ & Base & MS & $\Delta$ \\
\midrule
\multirow{3}{*}{\GemmaTwoTwoB}
  & 0 & 56.95 & 56.87 & $-$0.09 & 37.09 & 40.02 & $+$2.94 & 53.12 & 55.94 & $+$2.82 \\
  & 2 & 56.53 & 56.27 & $-$0.26 & 37.09 & 39.53 & $+$2.45 & 53.12 & 55.95 & $+$2.83 \\
  & 4 & 56.57 & 56.35 & $-$0.23 & 37.09 & 39.53 & $+$2.45 & 53.12 & 55.95 & $+$2.83 \\
\addlinespace
\multirow{3}{*}{\GemmaTwoNineB}
  & 0 & 71.88 & 71.92 & $+$0.04 & 42.84 & 42.96 & $+$0.12 & 60.15 & 60.70 & $+$0.55 \\
  & 2 & 72.33 & 72.03 & $-$0.30 & 42.84 & 42.96 & $+$0.12 & 60.15 & 60.65 & $+$0.49 \\
  & 4 & 72.17 & 72.01 & $-$0.16 & 42.84 & 42.96 & $+$0.12 & 60.15 & 60.75 & $+$0.59 \\
\addlinespace
\multirow{3}{*}{\LlamaThreeTwoThreeB}
  & 0 & 62.14 & 62.13 & $-$0.01 & 33.78 & 33.54 & $-$0.24 & 51.43 & 51.37 & $-$0.06 \\
  & 2 & 60.59 & 60.74 & $+$0.15 & 33.78 & 33.54 & $-$0.24 & 51.43 & 51.37 & $-$0.06 \\
  & 4 & 60.50 & 60.70 & $+$0.19 & 33.78 & 33.54 & $-$0.24 & 51.43 & 51.37 & $-$0.06 \\
\addlinespace
\multirow{3}{*}{\LlamaThreeOneEightB}
  & 0 & 68.34 & 68.34 & $+$0.01 & 38.19 & 37.94 & $-$0.24 & 54.51 & 54.71 & $+$0.20 \\
  & 2 & 68.17 & 68.08 & $-$0.09 & 38.19 & 37.94 & $-$0.24 & 54.51 & 54.71 & $+$0.19 \\
  & 4 & 68.34 & 68.32 & $-$0.01 & 38.19 & 37.94 & $-$0.24 & 54.51 & 54.79 & $+$0.28 \\
\addlinespace
\multirow{3}{*}{\QwenThreeZeroPointSixB}
  & 0 & 40.31 & 41.20 & $+$0.89 & 27.05 & 25.70 & $-$1.35 & 42.85 & 41.34 & $-$1.51 \\
  & 2 & 46.80 & 46.09 & $-$0.71 & 27.05 & 25.70 & $-$1.35 & 42.85 & 41.36 & $-$1.50 \\
  & 4 & 47.10 & 46.75 & $-$0.36 & 27.05 & 25.70 & $-$1.35 & 42.85 & 41.42 & $-$1.43 \\
\addlinespace
\multirow{3}{*}{\QwenThreeOnePointSevenB}
  & 0 & 55.50 & 54.97 & $-$0.53 & 29.74 & 29.38 & $-$0.37 & 46.00 & 46.21 & $+$0.21 \\
  & 2 & 59.19 & 58.99 & $-$0.21 & 29.74 & 29.38 & $-$0.37 & 46.00 & 46.19 & $+$0.19 \\
  & 4 & 59.83 & 59.78 & $-$0.06 & 29.74 & 29.38 & $-$0.37 & 46.00 & 46.34 & $+$0.34 \\
\addlinespace
\multirow{3}{*}{\QwenThreeFourB}
  & 0 & 72.48 & 71.93 & $-$0.55 & 43.75 & 43.75 & $+$0.00 & 65.44 & 67.60 & $+$2.16 \\
  & 2 & 72.26 & 72.04 & $-$0.22 & 43.75 & 43.75 & $+$0.00 & 65.44 & 66.73 & $+$1.29 \\
  & 4 & 72.81 & 72.81 & $+$0.00 & 43.75 & 43.75 & $+$0.00 & 65.44 & 66.38 & $+$0.94 \\
\bottomrule
\end{tabular}
\end{table*}

\section{Steering as a Geometric Transformation}
\label{app:steering_geometric_transformation}

The analysis presented in Appendix~\ref{app:cross_model_similarity} indicates that steering has a noisy influence on the generated text at the local level, particularly at the word level. However, when the generated text is represented at a coarser granularity---for example, at the sentence level using pooling---the overall structure of the response remains comparatively stable.

Figure~\ref{fig:three_trajectory_plots}\textbf{\captiona}--\textbf{\captionb} visualizes trajectories obtained from responses generated under different steering directions. Each trajectory is constructed from the first six points extracted from a generated response. To make the relationship between the trajectories and the generated language explicit, the instruction, steering direction, and complete recorded response are provided below for each example.

The examples show that steering changes the wording, tone, formatting, and organizational structure of the generated text. Nevertheless, the responses generally preserve the broader semantic objective specified by the instruction. Thus, steering appears to influence the local realization of the response while leaving several global properties of the trajectory---including its general curvature, velocity profile, and semantic direction---comparatively stable.

The following examples reproduce the recorded generated responses in full. Some supplied responses terminate mid-sentence; these endings are retained as recorded.

\subsection{Instruction: ``How to Lose Weight?''}

\subsubsection{Generated Response 1}

\noindent\textbf{Instruction:} \textit{How to lose weight?}

\medskip
\noindent\textbf{Steering direction:}
\textit{Particular patterns related to connectivity and relationships in a structured format} (\#3).

\medskip
\noindent\textbf{Final generated response:}

\begin{generatedresponse}
\responseitem{Establish clear goals within a network of support, such as friends or family engaged in similar health journeys.}

\responseitem{Create routines that interconnect meal planning and physical activity, ensuring a flow that promotes accountability.}

\responseitem{Monitor progress while maintaining relationships with health professionals, integrating their insights into a structure that is adaptable.}

\responseitem{Leverage community resources, joining group fitness or dieting programs to foster connections that reinforce commitment.}

\responseitem{Reflect on emotional patterns surrounding food choices, identifying connections between feelings, habits, and health objectives to achieve a balanced approach.}
\end{generatedresponse}

\subsubsection{Generated Response 2}

\noindent\textbf{Instruction:} \textit{How to lose weight?}

\medskip
\noindent\textbf{Steering direction:}
\textit{Entities, particularly names and classifications, as well as structured data elements} (\#4).

\medskip
\noindent\textbf{Final generated response:}

\begin{generatedresponse}
\responseitem{Nutritional categories: carbohydrates, proteins, fats.}

\responseitem{Activity types: aerobic, anaerobic, flexibility.}

\responseitem{Weight-loss entities: caloric deficit, metabolism, body composition.}

\responseitem{Structured data: weekly meal plan, daily exercise log, progress tracker.}

\responseitem{Classification examples.}
\end{generatedresponse}

\subsubsection{Generated Response 3}

\noindent\textbf{Instruction:} \textit{How to lose weight?}

\medskip
\noindent\textbf{Steering direction:}
\textit{Non-textual elements or formatting indicators present in a document} (\#5).

\medskip
\noindent\textbf{Final generated response:}

\begin{generatedresponse}
\begin{center}
\textbf{How to Lose Weight: A Visual Guide}
\end{center}

\responseitem{\textbf{Images:} Incorporate before-and-after photos to illustrate progress.}

\responseitem{\textbf{Charts:} Use bar graphs to show calorie intake versus calorie expenditure over time.}

\responseitem{\textbf{Icons:} Utilize food icons to represent healthy meal choices such as fruits and vegetables.}

\responseitem{\textbf{Tables:} Create a meal-planning table that outlines daily calorie goals and meal options.}

\responseitem{\textbf{Color Coding:} Apply different colors to categorize exercises, for example, cardio in blue and strength training in green.}

\responseitem{\textbf{.}}
\end{generatedresponse}

\subsubsection{Generated Response 4}

\noindent\textbf{Instruction:} \textit{How to lose weight?}

\medskip
\noindent\textbf{Steering direction:}
\textit{Elements related to lists and structured requests} (\#6).

\medskip
\noindent\textbf{Final generated response:}

\begin{generatedresponse}
\responseitem{Set clear goals.}

\responseitem{Track food intake.}

\responseitem{Incorporate regular exercise.}

\responseitem{Prioritize hydration.}

\responseitem{Focus on whole foods.}

\responseitem{Avoid sugary drinks.}

\responseitem{Plan meals in advance.}

\responseitem{Monitor portion sizes.}

\responseitem{Get enough sleep.}

\responseitem{Stay consistent with habits.}
\end{generatedresponse}

\subsubsection{Generated Response 5}

\noindent\textbf{Instruction:} \textit{How to lose weight?}

\medskip
\noindent\textbf{Steering direction:}
\textit{Procedural instructions related to software or game installation} (\#7).

\medskip
\noindent\textbf{Final generated response:}

\begin{generatedresponse}
\responseitem{Open the application interface for your weight-management program. Ensure that all prerequisites are met, including dietary preferences and exercise restrictions.}

\responseitem{Download the meal-plan module. Once the download is complete, execute the program to initiate the dietary-customization process.}

\responseitem{Navigate to the settings panel and adjust the caloric-intake parameters. Confirm your selections to ensure that the new limits are activated.}

\responseitem{Install the physical-activity tracker add-on. After installation, link it with your device for real-time updates on your exercises.}

\responseitem{Monitor progress through the analytics dashboard. Regularly check the updates and evaluate your metrics.}
\end{generatedresponse}

\subsection{Instruction: ``What Are Some Tools That Every Homeowner Should Have?''}

\noindent\textbf{Instruction:}
\textit{What are some tools that every homeowner should have?}

\subsubsection{Generated Response 1}

\noindent\textbf{Steering direction:}
\textit{Specific structural markers or formatting indicators in a document}.

\medskip
\noindent\textbf{Final generated response:}

\begin{generatedresponse}
\responseitem{\textbf{Hammer:} Essential for driving nails into or pulling them from materials, often marked by its head size and shape.}

\responseitem{\textbf{Screwdriver Set:} Featuring various tips, such as flat and Phillips, categorized for different screw types and crucial for assembly and repairs.}

\responseitem{\textbf{Tape Measure:} Typically sectioned in inches and centimeters, it provides accuracy in measuring spaces and objects and is indicated by clear markings.}

\responseitem{\textbf{Level:} Equipped with horizontal and vertical indicators, it ensures that surfaces are even and properly aligned, making it vital for installations.}

\responseitem{\textbf{Pliers:} Differentiated by type (needle-nose, slip-joint),.}
\end{generatedresponse}

\subsubsection{Generated Response 2}

\noindent\textbf{Instruction:}
\textit{What are some tools that every homeowner should have?}

\medskip
\noindent\textbf{Steering direction:}
\textit{Keywords and structural elements relevant to programming or technical documentation}.

\medskip
\noindent\textbf{Final generated response:}

\begin{generatedresponse}
\responseitem{\textbf{Hammer:} Essential for executing tasks that require impact and precision; serves as the primary execution framework for fastening components.}

\responseitem{\textbf{Screwdriver:} Integral for accessing or manipulating closures; supports various modules via interchangeable bits for different operations.}

\responseitem{\textbf{Pliers:} Provides gripping capabilities, enabling multi-use functionality similar to error handling; essential for enhancing torque and manipulating non-digital components.}

\responseitem{\textbf{Level:} A measurement tool essential for ensuring alignment; offers validation of horizontal and vertical parameters in project specifications.}

\responseitem{\textbf{Tape Measure:} Fundamental for quantifying distances; functions as a scoring mechanism to determine the size and.}
\end{generatedresponse}

\subsubsection{Generated Response 3}

\noindent\textbf{Instruction:}
\textit{What are some tools that every homeowner should have?}

\medskip
\noindent\textbf{Steering direction:}
\textit{Components related to job listings and their associated properties in a structured format}.

\medskip
\noindent\textbf{Final generated response:}

\begin{generatedresponse}
\responseitem{\textbf{Hammer:} Essential for various household tasks, akin to a key skill in job listings.}

\responseitem{\textbf{Screwdriver set:} A fundamental tool for assembly, similar to the necessary qualifications for a position.}

\responseitem{\textbf{Tape measure:} Vital for accurate measurements, paralleling the precision required in job descriptions.}

\responseitem{\textbf{Level:} Ensures that projects are even, much like maintaining balance in job responsibilities.}

\responseitem{\textbf{Pliers:} Useful for gripping and twisting, reflecting the adaptability sought in candidates.}

\responseitem{\textbf{Utility knife:} Versatile for cutting tasks, resembling multi-talented applicants in a competitive market.}

\responseitem{\textbf{Wrench set:} Critical for.}
\end{generatedresponse}

\subsubsection{Generated Response 4}

\noindent\textbf{Instruction:}
\textit{What are some tools that every homeowner should have?}

\medskip
\noindent\textbf{Steering direction:}
\textit{Terms related to articles and forms in a structured format}.

\medskip
\noindent\textbf{Final generated response:}

\begin{generatedresponse}
\responseitem{\textbf{Hammer:} Essential tool, typically categorized under fasteners and attachments.}

\responseitem{\textbf{Screwdriver:} Commonly found in home-maintenance checklists and classified as a specific tool.}

\responseitem{\textbf{Wrench:} Integral for plumbing tasks and noted within repair documentation.}

\responseitem{\textbf{Pliers:} Frequently referenced in do-it-yourself guides and serving various functions in toolboxes.}

\responseitem{\textbf{Tape Measure:} Critical for measurements and usually documented in project-planning forms.}

\responseitem{\textbf{Level:} Important for ensuring alignment and often included in construction articles.}

\responseitem{\textbf{Utility Knife:} Listed in safety protocols and widely used for cutting tasks.}

\responseitem{\textbf{Drill:} Noted in inventory lists,.}
\end{generatedresponse}

\colorlet{I1}{red!75!black}       \colorlet{I1f}{red!60!white}
\colorlet{I2}{orange!85!black}    \colorlet{I2f}{orange!60!white}
\colorlet{I3}{yellow!55!black}    \colorlet{I3f}{yellow!55!white}
\colorlet{I4}{green!65!black}     \colorlet{I4f}{green!55!white}
\colorlet{I5}{teal!75!black}      \colorlet{I5f}{teal!55!white}
\colorlet{I6}{cyan!75!black}      \colorlet{I6f}{cyan!55!white}
\colorlet{I7}{blue!75!black}      \colorlet{I7f}{blue!55!white}
\colorlet{I8}{violet!75!black}    \colorlet{I8f}{violet!55!white}
\colorlet{I9}{magenta!75!black}   \colorlet{I9f}{magenta!55!white}
\colorlet{I10}{brown!75!black}    \colorlet{I10f}{brown!55!white}
\colorlet{I11}{olive!75!black}    \colorlet{I11f}{olive!55!white}

\newtcolorbox{conceptlist}[1][]{
    breakable=false,
    colback=cyan!8,
    colframe=cyan!45!blue,
    boxrule=0.4pt,
    arc=2pt,
    left=6pt,
    right=6pt,
    top=5pt,
    bottom=5pt,
    before skip=6pt,
    after skip=10pt,
    #1
}

\newenvironment{concepttable}
{%
    \small
    \renewcommand{\arraystretch}{1.15}
    \begin{tabular}{@{}p{1.35cm}p{\dimexpr\linewidth-1.35cm-4\tabcolsep\relax}@{}}
    \toprule
    \textbf{ID} & \textbf{Concept} \\
    \midrule
}
{%
    \bottomrule
    \end{tabular}
}

\begin{table}[t]
    \centering
    \small
    \setlength{\tabcolsep}{3pt}
    \caption{Inter-rater agreement across six model-based raters and 150 evaluation cases.}
    \label{tab:inter_rater_agreement}
    \begin{tabular}{@{}l@{\hspace{5pt}}r@{\hspace{5pt}}p{0.43\columnwidth}@{}}
        \toprule
        \textbf{Metric} & \textbf{Value} & \textbf{Notes} \\
        \midrule

        \multicolumn{3}{@{}l}{\textit{Intraclass correlation}} \\
        ICC(2,1) & 0.396 & Single-rater absolute agreement \\
        ICC(2,k) & 0.798 & Average of six raters under absolute agreement \\
        ICC(3,1) & 0.408 & Single-rater consistency \\
        ICC(3,k) & 0.805 & Average of six raters under consistency \\[2pt]

        \multicolumn{3}{@{}l}{\textit{Rank-based agreement}} \\
        Kendall's $W$ & 0.441 &
        $\chi^2(149)=394.3$, $p=4.6\times 10^{-24}$ \\
        Mean pairwise Spearman $\rho$ & 0.329 &
        Mean Spearman correlation implied by $W$ \\

        \bottomrule
    \end{tabular}
\end{table}

\section{Concept Inventory and Inter-Rater Agreement}
\label{app:concept_inventory}

The experimental details of the concept-level geometric analysis---including
concept selection, construction of the CAA and MetaSteer directions, cosine
similarity computation, and model evaluation---are provided in
Appendix~\ref{app:concept_linearity}. This appendix documents the concept
inventory and reports inter-rater agreement for the associated evaluations.

For each concept, six model-based raters assessed the corresponding steering
behavior. The raters were drawn from the Llama, Gemma, and Qwen model
families. The resulting inter-rater agreement statistics are reported in
Table~\ref{tab:inter_rater_agreement}.

The results indicate moderate single-rater reliability, with an intraclass
correlation of approximately $0.40$. Agreement increases substantially when
ratings are averaged across the six raters, reaching approximately $0.80$.
Kendall's $W$ and the mean pairwise Spearman correlation likewise indicate
moderate but statistically significant rank agreement.

The complete concept inventory and cluster assignments are provided below.
Cluster names are kept outside the shaded lists, while the concepts are
grouped within compact colored boxes.

\subsection{Concept Inventory by Cluster}
\label{app:concepts_by_cluster}

\subsubsection*{Response language}

\begin{conceptlist}[colback=I1f, colframe=I1]
\begin{concepttable}
5 & Response in Korean \\
3 & Response in Chinese \\
4 & Response in Japanese \\
\end{concepttable}
\end{conceptlist}

\subsubsection*{Format markers}

\begin{conceptlist}[colback=I2f, colframe=I2]
\begin{concepttable}
0 & Response with emojis \\
28 & Response with a YAML block \\
41 & Response contains detectable sarcasm markers \\
44 & Response uses a consistently informal tone \\
103 & Response mentions literature or literary analysis \\
48 & Response expresses high enthusiasm, such as through exclamations or positive framing \\
36 & Response uses a checklist or task-list format \\
27 & Response contains a JSON object \\
12 & Response is written in a single paragraph \\
\end{concepttable}
\end{conceptlist}

\subsubsection*{Persona \& tone}

\begin{conceptlist}[colback=I3f, colframe=I3]
\begin{concepttable}
141 & Response is written in the persona of a product designer, with a user-centered and UX-focused perspective \\
89 & Response is written in the persona of a friendly peer, using a casual and collaborative style \\
122 & Response opens with a direct greeting such as ``Hi!'', ``Hello!'', or ``Hey!'' \\
8 & Response is written in iambic-like poetic meter \\
75 & Response contains frequent metaphors \\
113 & Response is organized with clear sections and subheadings \\
143 & Response is written in the persona of a startup founder, emphasizing vision, speed, and iteration \\
45 & Response contains explicit polite markers such as ``please'' or ``thank you'' \\
15 & Response contains a Markdown table \\
142 & Response is written in the persona of a policy analyst, emphasizing trade-offs, stakeholders, and impact \\
50 & Response contains explicit empathetic language \\
55 & Response begins with an explicit outline or plan \\
125 & Response contains a rhetorical device \\
\end{concepttable}
\end{conceptlist}

\subsubsection*{Content devices}

\begin{conceptlist}[colback=I4f, colframe=I4]
\begin{concepttable}
56 & Response provides step-by-step instructions \\
33 & Response contains at least one emoji per sentence \\
40 & Response contains at least one clear joke or punchline \\
87 & Response is written in the persona of a motivational coach, using an encouraging and action-oriented style \\
16 & Response uses Markdown headings \\
99 & Response mentions cooking or recipes \\
146 & Response includes mathematical reasoning \\
29 & Response includes a regular-expression pattern example \\
105 & Response is written as a case study with a concrete scenario \\
140 & Response mentions music \\
22 & Response contains many exclamation marks \\
38 & Response uses repeated first letters, such as alliteration, in a sentence \\
119 & Response emphasizes novel or creative ideas \\
43 & Response uses a consistently formal tone \\
91 & Response is written in the persona of a storyteller, using a narrative-driven style \\
112 & Response is written as a comprehensive report \\
\end{concepttable}
\end{conceptlist}

\subsubsection*{Expert framing}

\begin{conceptlist}[colback=I5f, colframe=I5]
\begin{concepttable}
66 & Response explains a concept using analogies \\
145 & Response mentions climate change \\
62 & Response frames the answer around risk or safety \\
77 & Response uses technical jargon in an expert explanation \\
98 & Response mentions computer programming \\
81 & Response is written in the persona of a research scientist, using a technical and evidence-based style \\
130 & Response uses softening language such as ``I suggest'' or ``it may help'' \\
84 & Response is written in the persona of a software engineer, using a practical and implementation-focused style \\
147 & Response is written in the style of a Twitter post \\
52 & Response includes explicit uncertainty disclaimers \\
80 & Response is written in the persona of a teacher, using an explanatory, structured, and pedagogical style \\
69 & Response contains a ``Common Pitfalls'' section \\
83 & Response is written in the persona of a doctor, using a careful, qualified, and safety-aware style \\
110 & Response explicitly weighs pros and cons in decision-making \\
95 & Response mentions machine learning \\
42 & Response follows a scientific writing style that is neutral, precise, and impersonal \\
\end{concepttable}
\end{conceptlist}

\subsubsection*{Explanatory structure}

\begin{conceptlist}[colback=I6f, colframe=I6]
\begin{concepttable}
53 & Response asks at least one clarifying question \\
102 & Response mentions health or fitness \\
82 & Response is written in the persona of a lawyer, using a precise, conditional, and cautious style \\
78 & Response uses ``we'' or ``our'' at least twice to frame a collaborative perspective \\
106 & Response is written as a textbook-style explanation \\
115 & Response frames outcomes in terms of opportunities \\
120 & Response contains at least one date expression \\
79 & Response includes a short ``Next steps:'' section with actionable items \\
71 & Response provides a minimal working example \\
116 & Response frames outcomes in terms of risks or downsides \\
92 & Response is written in the persona of a consultant, using a structured and actionable style \\
144 & Response is written in the persona of a technical writer, emphasizing clarity and documentation \\
100 & Response mentions travel planning \\
149 & Response mentions a specific named country \\
118 & Response grounds claims with explicit evidence or references \\
132 & Response includes at least one rhetorical question \\
\end{concepttable}
\end{conceptlist}

\subsubsection*{List \& decision format}

\begin{conceptlist}[colback=I7f, colframe=I7]
\begin{concepttable}
11 & Response is presented as a numbered list \\
136 & Response uses an example-first structure followed by an explanation \\
63 & Response frames the answer around monetary cost \\
94 & Response is written in the persona of a tutor, using guided and stepwise explanations \\
31 & Response includes a short title line \\
96 & Response mentions mathematics \\
67 & Response explains a concept using counterexamples \\
148 & Response is framed as a risk-benefit analysis \\
47 & Response contains frequent hedging words such as ``might'', ``maybe'', or ``likely'' \\
10 & Response contains exactly five bullet points \\
1 & Response is written in uppercase \\
107 & Response is written as a brief executive summary \\
25 & Response contains the phrase ``it is worth noting'' \\
61 & Response frames the answer around efficiency or optimization \\
65 & Response frames the answer around privacy concerns \\
32 & Response uses only ASCII characters \\
\end{concepttable}
\end{conceptlist}

\subsubsection*{Opening/closing conventions}

\begin{conceptlist}[colback=I8f, colframe=I8]
\begin{concepttable}
26 & Response opens with ``Of course'' \\
19 & Response uses bold emphasis \\
34 & Response ends with a sentence beginning with ``In conclusion'' \\
74 & Response is written as a question-and-answer dialogue \\
101 & Response mentions finance or investing \\
7 & Response begins with the phrase ``Let me'' \\
135 & Response uses passive voice at least once, such as ``it is said'' or ``this can be done'' \\
121 & Response begins with a question \\
23 & Response ends with a question \\
86 & Response is written in the persona of a strict grader, using a critical and rubric-driven style \\
30 & Response contains at least one equation in LaTeX \\
\end{concepttable}
\end{conceptlist}

\subsubsection*{Answer templates}

\begin{conceptlist}[colback=I9f, colframe=I9]
\begin{concepttable}
59 & Response contains an explicit warning or caution \\
68 & Response begins with a formal definition \\
6 & Response is written in rhyming couplets \\
20 & Response uses italic emphasis \\
14 & Response contains exactly four sentences \\
108 & Response is written as a frequently asked questions (FAQ) answer \\
9 & Response contains exactly three bullet points \\
111 & Response is written in a headline-style format \\
54 & Response includes a brief self-check or sanity check \\
64 & Response frames the answer around time or latency \\
139 & Response uses conditional reasoning, such as ``if \ldots then \ldots'' \\
123 & Response contains at least one time expression \\
35 & Response addresses the user with ``you'' or ``your'' at least three times \\
126 & Response begins by approving the user's question, such as ``That's a great question!'' \\
60 & Response states its assumptions explicitly \\
97 & Response mentions physics \\
21 & Response contains no punctuation \\
24 & Response contains the phrase ``in other words'' at least once \\
137 & Response uses an explanation-first structure followed by an example \\
85 & Response is written in the persona of a customer-support agent, using a polite and problem-solving style \\
51 & Response uses assertive and confident phrasing \\
\end{concepttable}
\end{conceptlist}

\subsubsection*{Discourse connectors}

\begin{conceptlist}[colback=I10f, colframe=I10]
\begin{concepttable}
70 & Response includes a short quiz-style question \\
76 & Response avoids technical jargon and uses a lay explanation \\
127 & Response uses ``also'' or ``additionally'' to introduce at least two separate points \\
129 & Response ends with a closing sentence beginning with ``In short'' \\
18 & Response ends with the phrase ``Let me know if you have any questions'' \\
128 & Response uses strong imperative verbs, such as ``Do X'' or ``Avoid Y'' \\
2 & Response is written in lowercase \\
57 & Response provides multiple alternative options \\
46 & Response avoids hedging words such as ``might'', ``maybe'', or ``likely'' \\
37 & Response uses the word ``because'' at least twice to explain its reasoning \\
58 & Response gives a single clear recommendation \\
104 & Response mentions history \\
90 & Response is written in the persona of a journalist, using a neutral and fact-focused style \\
49 & Response expresses skepticism or doubt \\
72 & Response lists exactly two supporting reasons \\
\end{concepttable}
\end{conceptlist}

\subsubsection*{Argumentative framing}

\begin{conceptlist}[colback=I11f, colframe=I11]
\begin{concepttable}
114 & Response is written as a brainstorming-style answer \\
131 & Response uses a consistently decisive tone \\
117 & Response frames outcomes in a neutral and descriptive way \\
13 & Response contains exactly two sentences \\
138 & Response repeatedly references the user's goal or intent \\
17 & Response uses the word ``key'' at least twice \\
39 & Response includes at least one hyperlink using an \texttt{http} or \texttt{https} address \\
124 & Response contains at least one parenthetical remark \\
109 & Response is written as a point--counterpoint argument \\
88 & Response is written in the persona of a skeptical reviewer, using a critical and evidence-demanding style \\
134 & Response explicitly acknowledges the user's goal \\
133 & Response uses contrastive markers such as ``however'' or ``on the other hand'' \\
93 & Response is written in the persona of a debate opponent, using an argumentative and contrastive style \\
73 & Response explicitly contrasts two viewpoints \\
\end{concepttable}
\end{conceptlist}

\section{Cross-Model Similarity of Steering Trajectories}
\label{app:cross_model_similarity}

We examine whether responses to the same instruction retain common geometric
structure across steering concepts, model families, and encoding modes.
Following \citet{zhou2026geometry}, we compare position, velocity,
acceleration, and Menger-curvature representations of hidden-state
trajectories. We additionally use sentence-order shuffling as a negative
control for ordered trajectory structure.

\paragraph{Data and models.}

We combine Concept16K-v1 and Concept16K-v2 \ref{app:axbench_dataset} into a dataset of 131,363 records,
covering 2,134 instructions, 10,488 steering concepts, and three response
genres: \texttt{code}, \texttt{text}, and \texttt{math}. To reduce the
influence of sparsely represented groups, we retain only records whose
instruction, concept, and genre each occur more than 50 times. This produces
2,550 instruction--concept pairs spanning 381 instructions, 70 concepts, and
all three genres.

We evaluate the six primary backbones and additionally include \QwenThreeZeroPointSixB
as a small-model capability-retention stress test.
\QwenThreeOnePointSevenB, \QwenThreeFourB, \GemmaTwoTwoB,
\GemmaTwoNineB, \LlamaThreeTwoThreeB, and \LlamaThreeOneEightB.
Responses are represented using the steps supplied by the dataset; responses
stored as single strings are segmented at newlines or sentence-ending
punctuation. The retained responses contain 29,484 segments in total.

\paragraph{Cumulative and isolated representations.}

Let $x_1,\ldots,x_T$ be the ordered segments of a response and let
\[
S_t=x_1\Vert\cdots\Vert x_t
\]
denote the cumulative prefix. We extract representations under two encoding
modes, $m\in\{\mathrm{cum},\mathrm{iso}\}$.

In cumulative mode, segment $x_t$ is encoded with all preceding segments:
\begin{equation}
\mathbf{s}^{(\mathrm{cum})}_t
=
\frac{1}{|I_t|}
\sum_{j\in I_t}
\mathbf{h}^{(L)}_j(S_t),
\label{eq:cumulative-representation}
\end{equation}
where $I_t$ contains the token positions introduced by $x_t$.

In isolated mode, each segment is encoded independently:
\begin{equation}
\mathbf{s}^{(\mathrm{iso})}_t
=
\frac{1}{|I_t|}
\sum_{j\in I_t}
\mathbf{h}^{(L)}_j(x_t).
\label{eq:isolated-representation}
\end{equation}
Thus, both modes use the same segments and pooling rule, but only cumulative
encoding allows preceding segments to contextualize the current segment.

For either mode, we define position, velocity, and acceleration vectors as
\begin{align}
\mathbf{p}^{(m)}_t
&=
\mathbf{s}^{(m)}_t,
\\
\mathbf{v}^{(m)}_t
&=
\mathbf{s}^{(m)}_{t+1}-\mathbf{s}^{(m)}_t,
\\
\mathbf{a}^{(m)}_t
&=
\mathbf{v}^{(m)}_{t+1}-\mathbf{v}^{(m)}_t.
\end{align}
The scalar quantity used as speed in the main text is
\begin{equation}
v^{(m)}_t=\|\mathbf{v}^{(m)}_t\|_2.
\end{equation}
Our similarity analysis retains the complete vector
$\mathbf{v}^{(m)}_t$, rather than only its magnitude.

Menger curvature is computed from three consecutive states:
\begin{equation}
\kappa^{(m)}_t
=
\frac{
4\,\operatorname{Area}
\left(
\mathbf{s}^{(m)}_{t-1},
\mathbf{s}^{(m)}_t,
\mathbf{s}^{(m)}_{t+1}
\right)
}{
\|\mathbf{s}^{(m)}_{t-1}-\mathbf{s}^{(m)}_t\|_2
\|\mathbf{s}^{(m)}_t-\mathbf{s}^{(m)}_{t+1}\|_2
\|\mathbf{s}^{(m)}_{t+1}-\mathbf{s}^{(m)}_{t-1}\|_2
}.
\label{eq:appendix-curvature}
\end{equation}

\paragraph{Trajectory similarity.}

For $r\in\{\mathbf{p},\mathbf{v},\mathbf{a}\}$, the similarity between
trajectories $i$ and $j$ is the mean cosine similarity between temporally
aligned vectors:
\begin{equation}
s_{ij}^{(r,m)}
=
\frac{1}{\ell_{ij}^{(r,m)}}
\sum_{t=1}^{\ell_{ij}^{(r,m)}}
\frac{
\left\langle
\mathbf{r}^{(m)}_{i,t},
\mathbf{r}^{(m)}_{j,t}
\right\rangle
}{
\|\mathbf{r}^{(m)}_{i,t}\|_2
\|\mathbf{r}^{(m)}_{j,t}\|_2
},
\label{eq:trajectory-similarity}
\end{equation}
where $\ell_{ij}^{(r,m)}$ is the shorter sequence length. Menger-curvature
similarity is the Pearson correlation between aligned scalar sequences:
\begin{equation}
s_{ij}^{(\kappa,m)}
=
\operatorname{Corr}
\left(
\kappa^{(m)}_{i,1:\ell},
\kappa^{(m)}_{j,1:\ell}
\right).
\label{eq:curvature-similarity}
\end{equation}

We average similarities within instruction, steering-concept, and genre
groups. In Tables~\ref{tab:cumulative_similarity} and
\ref{tab:isolated_similarity}, \textbf{I}, \textbf{C}, and \textbf{G}
denote these respective grouping criteria. Each entry reports macro/micro
similarity. Macro averaging weights groups equally, whereas micro averaging
pools all within-group pairs.

\paragraph{Shuffled control.}

For every response, we independently permute its segment order using a fixed
random seed of 42. The same permutations are used across models. In cumulative
mode, the shuffled order changes both segment adjacency and the context
preceding each segment. In isolated mode, individual segment embeddings remain
context-free, but shuffling changes which segments are adjacent when velocity,
acceleration, and curvature are calculated.

\begin{table*}[t]
\centering
\caption{
Cumulative-context trajectory similarity. Each model is evaluated using its
original and shuffled segment order. Entries report macro/micro averages.
Position, velocity, and acceleration use mean cosine similarity; Menger
curvature uses Pearson correlation.
}
\label{tab:cumulative_similarity}
\resizebox{\textwidth}{!}{
\begin{tabular}{ll*{12}{c}}
\toprule
&& \multicolumn{3}{c}{Position}
& \multicolumn{3}{c}{Velocity}
& \multicolumn{3}{c}{Acceleration}
& \multicolumn{3}{c}{Menger curvature}\\
\cmidrule(lr){3-5}
\cmidrule(lr){6-8}
\cmidrule(lr){9-11}
\cmidrule(lr){12-14}
\textbf{Model} & \textbf{Order}
& \textbf{I} & \textbf{C} & \textbf{G}
& \textbf{I} & \textbf{C} & \textbf{G}
& \textbf{I} & \textbf{C} & \textbf{G}
& \textbf{I} & \textbf{C} & \textbf{G}\\
\midrule
\QwenThreeOnePointSevenB & Original
& .886/.908 & .842/.810 & .861/.802
& .308/.386 & .154/.104 & .109/.092
& .271/.349 & .119/.068 & .075/.058
& .277/.349 & .116/.041 & .111/.032\\
& Shuffled
& .927/.936 & .904/.903 & .912/.900
& .043/.045 & .041/.038 & .050/.038
& .015/.017 & .013/.013 & .019/.013
& .115/.122 & .055/.113 & .091/.112\\
\addlinespace

\QwenThreeFourB & Original
& .813/.852 & .748/.698 & .771/.682
& .306/.385 & .145/.106 & .097/.094
& .270/.350 & .112/.069 & .060/.058
& .256/.350 & .137/.041 & .134/.031\\
& Shuffled
& .840/.860 & .810/.783 & .834/.773
& .048/.050 & .043/.040 & .050/.039
& .020/.020 & .016/.015 & .022/.014
& .170/.168 & .108/.149 & .128/.148\\
\addlinespace

\GemmaTwoTwoB & Original
& .939/.954 & .913/.915 & .904/.907
& .305/.379 & .174/.108 & .118/.096
& .271/.346 & .145/.075 & .086/.064
& .377/.453 & .296/.189 & .324/.182\\
& Shuffled
& .931/.942 & .912/.919 & .905/.914
& .070/.070 & .059/.062 & .067/.061
& .038/.035 & .028/.031 & .040/.031
& .364/.421 & .277/.409 & .262/.400\\
\addlinespace

\GemmaTwoNineB & Original
& .757/.801 & .669/.625 & .667/.610
& .300/.373 & .162/.112 & .121/.102
& .261/.336 & .130/.074 & .083/.064
& .536/.600 & .529/.381 & .562/.380\\
& Shuffled
& .704/.725 & .659/.633 & .668/.621
& .073/.074 & .064/.065 & .065/.065
& .031/.031 & .029/.027 & .031/.026
& .590/.616 & .582/.594 & .641/.587\\
\addlinespace

\LlamaThreeTwoThreeB & Original
& .742/.795 & .638/.563 & .656/.543
& .302/.379 & .149/.099 & .102/.087
& .271/.351 & .124/.073 & .073/.062
& .313/.396 & .217/.082 & .226/.075\\
& Shuffled
& .708/.739 & .653/.589 & .680/.573
& .047/.050 & .043/.038 & .045/.037
& .020/.020 & .021/.015 & .023/.015
& .263/.284 & .208/.265 & .209/.265\\
\addlinespace

\LlamaThreeOneEightB & Original
& .780/.830 & .654/.628 & .657/.609
& .306/.381 & .147/.106 & .100/.094
& .272/.350 & .115/.075 & .065/.063
& .406/.486 & .372/.227 & .344/.223\\
& Shuffled
& .738/.769 & .654/.612 & .671/.598
& .062/.065 & .059/.052 & .060/.051
& .023/.023 & .023/.018 & .025/.017
& .444/.471 & .364/.460 & .373/.460\\
\midrule
\multicolumn{2}{l}{\textbf{Original mean}}
& .819/.857 & .744/.707 & .752/.692
& \textbf{.305/.381 }& .155/.106 & .108/.094
& \textbf{.269/.347} & .124/.072 & .074/.061
& .361/.439 & .278/.160 & .283/.154\\
\multicolumn{2}{l}{\textbf{Shuffled mean}}
& .808/.829 & .765/.740 & .778/.730
& .057/.059 & .051/.049 & .056/.048
& .024/.024 & .022/.020 & .027/.019
& .324/.347 & .266/.331 & .284/.329\\
\bottomrule
\end{tabular}
}
\end{table*}

\begin{table*}[t]
\centering
\caption{
Isolated-segment trajectory similarity. Each segment is encoded without
preceding context. Shuffling therefore changes segment adjacency but not the
context used to embed an individual segment. Entries report macro/micro
averages.
}
\label{tab:isolated_similarity}
\resizebox{\textwidth}{!}{
\begin{tabular}{ll*{12}{c}}
\toprule
&& \multicolumn{3}{c}{Position}
& \multicolumn{3}{c}{Velocity}
& \multicolumn{3}{c}{Acceleration}
& \multicolumn{3}{c}{Menger curvature}\\
\cmidrule(lr){3-5}
\cmidrule(lr){6-8}
\cmidrule(lr){9-11}
\cmidrule(lr){12-14}
\textbf{Model} & \textbf{Order}
& \textbf{I} & \textbf{C} & \textbf{G}
& \textbf{I} & \textbf{C} & \textbf{G}
& \textbf{I} & \textbf{C} & \textbf{G}
& \textbf{I} & \textbf{C} & \textbf{G}\\
\midrule
\QwenThreeOnePointSevenB & Original
& .957/.967 & .947/.938 & .949/.935
& .280/.362 & .133/.075 & .072/.061
& .260/.345 & .115/.052 & .048/.040
& .308/.405 & .070/.096 & .048/.083\\
& Shuffled
& .939/.946 & .940/.932 & .945/.930
& .001/.001 & -.004/.000 & .007/.000
& .004/.004 & .000/.002 & .013/.001
& .008/.003 & -.007/.000 & .006/.000\\
\addlinespace

\QwenThreeFourB & Original
& .887/.912 & .872/.840 & .880/.830
& .278/.358 & .129/.066 & .068/.053
& .257/.340 & .113/.047 & .049/.035
& .265/.359 & .040/.044 & .028/.033\\
& Shuffled
& .847/.863 & .855/.827 & .870/.819
& .002/.001 & -.001/.000 & .006/.000
& .004/.002 & -.001/.001 & .011/.000
& .018/.002 & .001/.000 & .003/.001\\
\addlinespace

\GemmaTwoTwoB & Original
& .945/.956 & .930/.932 & .926/.927
& .312/.384 & .180/.103 & .131/.090
& .292/.366 & .162/.085 & .107/.073
& .300/.372 & .047/.030 & .057/.022\\
& Shuffled
& .924/.932 & .919/.924 & .915/.921
& .002/.000 & .000/.001 & .005/.000
& .003/.001 & -.002/.001 & .008/.000
& .011/.007 & .007/.001 & -.003/.001\\
\addlinespace

\GemmaTwoNineB & Original
& .878/.896 & .860/.826 & .855/.822
& .314/.386 & .203/.101 & .155/.089
& .295/.367 & .188/.084 & .138/.074
& .283/.362 & .154/.028 & .121/.021\\
& Shuffled
& .823/.830 & .829/.806 & .828/.804
& .001/.001 & .001/.000 & .004/.000
& .000/.001 & .000/.001 & .007/.000
& .015/-.004 & -.015/.000 & -.003/.000\\
\addlinespace

\LlamaThreeTwoThreeB & Original
& .762/.803 & .723/.643 & .741/.630
& .278/.353 & .126/.061 & .072/.049
& .257/.334 & .110/.044 & .055/.034
& .287/.370 & .054/.040 & .033/.030\\
& Shuffled
& .669/.690 & .684/.612 & .715/.602
& .001/.001 & .003/.000 & .005/.000
& .002/.002 & .004/.001 & .008/.000
& .030/.009 & .004/.002 & .003/.001\\
\addlinespace

\LlamaThreeOneEightB & Original
& .756/.800 & .691/.622 & .693/.610
& .273/.350 & .121/.058 & .060/.045
& .253/.332 & .105/.041 & .041/.030
& .291/.368 & .067/.053 & .046/.043\\
& Shuffled
& .661/.687 & .657/.591 & .671/.583
& .001/.001 & .005/.000 & .006/.000
& .001/.002 & .007/.001 & .009/.000
& .025/.008 & .001/.000 & .007/.001\\
\midrule
\multicolumn{2}{l}{\textbf{Original mean}}
& .864/.889 & .837/.800 & .841/.792
& \textbf{.289/.365} & .149/.077 & .093/.065
& \textbf{.269/.347} & .132/.059 & .073/.047
& \textbf{.289/.373} & .072/.049 & .056/.038\\
\multicolumn{2}{l}{\textbf{Shuffled mean}}
& .811/.825 & .814/.782 & .824/.776
& .001/.001 & .001/.000 & .005/.000
& .002/.002 & .001/.001 & .009/.000
& .018/.004 & -.002/.000 & .002/.001\\
\bottomrule
\end{tabular}
}
\end{table*}

\paragraph{Results.}
As shown in Tables~\ref{tab:cumulative_similarity} and
\ref{tab:isolated_similarity}, position similarity changes only modestly from
the shuffled to the original ordering and remains high under instruction,
concept, and genre grouping. Position similarity is therefore weakly
discriminative of ordered or instruction-specific structure; this does not
imply that positional representations contain no semantic information, but
rather that similarity at this level cannot reliably isolate it. In contrast,
instruction-grouped velocity exhibits substantial increases over the shuffled
baseline, from $.057/.059$ to $.305/.381$ under cumulative encoding and from
$.001/.001$ to $.289/.365$ under isolated encoding. Discrete acceleration
shows the same pattern, increasing from $.024/.024$ to $.269/.347$ in
cumulative mode and from $.002/.002$ to $.269/.347$ in isolated mode. These
results indicate that responses sharing an instruction possess meaningfully
aligned velocity and acceleration structure, and that this alignment depends
on the original segment ordering. Menger-curvature similarity likewise
increases strongly under isolated encoding, from $.018/.004$ to $.289/.373$.
The corresponding cumulative increase is weaker, from $.324/.347$ to
$.361/.439$, consistent with cumulative context inducing shared curvature
structure even after the segment order is shuffled.

\paragraph{Limitations.}

Models from the same family are related and should not be interpreted as
independent population samples. Instruction-grouped averages also include all
within-instruction pairs and do not require every pair to differ in both
concept and genre. The shuffled control destroys response coherence and is
therefore a negative control rather than a realistic steering condition.
Finally, Menger curvature is a scalar correlation-based measure and is more
sensitive to short trajectories and local irregularities than the vector-based
cosine similarities.

\section{Concept-Level Geometric Analysis}
\label{app:concept_linearity}

For each of the 150 concepts of \citet{fan2026your}, we compare two steering directions in the last-layer hidden state. CAA supplies a linear contrastive direction, the difference of mean last-prompt states on positive and negative examples,
\[
\mathbf{v}^{\mathrm{CAA}}_{c,m}
=
\mathbb{E}[h_m(x^{+})]-\mathbb{E}[h_m(x^{-})].
\]
MetaSteer supplies a more efficient intervention based on the results. Its effect can be partly described as displacement of the prompt state when the concept is applied,
\[
\mathbf{d}^{\mathrm{MS}}_{c,m}
=
\mathbb{E}_{i}\bigl[h_m(i,c)-h_m(i)\bigr].
\]
The share of that MetaSteer displacement captured by the CAA direction is their cosine,
\[
s_{c,m}
=
\cos\bigl(\mathbf{v}^{\mathrm{CAA}}_{c,m},\,\mathbf{d}^{\mathrm{MS}}_{c,m}\bigr).
\]

This cosine measures directional agreement between the CAA displacement and
the MetaSteer displacement. It is not an effective-rank measure and does not
estimate the intrinsic dimensionality of the hidden-state trajectory.
A value near one means the efficient displacement already lies along the linear CAA axis.

We compute \(s_{c,m}\) for every concept on six models: \GemmaTwoTwoB, \GemmaTwoNineB, \LlamaThreeTwoThreeB, \LlamaThreeOneEightB, \QwenThreeOnePointSevenB, and \QwenThreeFourB. The plotted concept CAA-alignment score is the mean standard deviation of this value over the 150 concepts. Figure~\ref{fig:three_trajectory_plots} ranks concepts by this capture score.

For each concept $c$, the plotted score is the mean of $s_{c,m}$ across the
six models, with shading showing one standard deviation across models.
Concepts with high scores have MetaSteer displacements aligned with the CAA
direction; low scores indicate disagreement with that particular direction.

\section{DPO Hyperparameter Configuration}
\label{app:dpo_details}
We performed preference optimization using Direct Preference Optimization (DPO) on top of a pretrained causal language model, with parameter-efficient fine-tuning through LoRA. In general, our pipeline used prompt--response preference pairs consisting of a preferred (\texttt{chosen}) and dispreferred (\texttt{rejected}) answer, and optimized the model to increase the likelihood of preferred responses relative to rejected ones. LoRA adapters were applied to the attention projection layers, with target modules \texttt{q\_proj}, \texttt{k\_proj}, \texttt{v\_proj}, and \texttt{o\_proj}, allowing efficient adaptation with reduced memory cost. Optimization was performed with the default AdamW-style optimizer from the training framework, using a learning rate of \texttt{1e-5}, weight decay of \texttt{0.01}, and a warmup ratio of \texttt{0.03}. The DPO temperature/preference strength parameter $\beta$ was set to \texttt{0.1}.

\section{Safety Evaluation Details}
\label{app:safety}

We use the 100 harmful behaviours of JailbreakBench~\citep{chao2024jailbreakbench}.
Because MetaSteer is conditioned on a separate concept $c$, a harmful request
can be decomposed so that the instruction itself is benign and the harmful
intent is carried entirely by the steering direction. Each behaviour was
split by an LLM into a benign instruction $x$ and a risky concept $c$, e.g.\
\emph{``Write a tutorial on how to make a bomb''} $\rightarrow$ $x=$
\emph{``tell me something useful''}, $c=$ \emph{``bomb manufacturing
recipe''}; or \emph{``Create a social media campaign that promotes body
shaming \ldots''} $\rightarrow$ $x=$ \emph{``write a creative social media
campaign for educational purposes''}, $c=$ \emph{``pretend to body shame
overweight people as a satire smartness test''}. No jailbreak templates,
adversarial suffixes, or optimisation against the model are used; the only
manipulation is the placement of the harmful intent in $c$.

Each $(x,c)$ pair is rendered with the training template of
Appendix~\ref{app:axbench_dataset} (\texttt{\#\#\# Instruction} /
\texttt{\#\#\# Steering Direction}) together with the system prompt ``You are
a helpful assistant.'', and is answered by the final MetaSteer checkpoint of
each model. Decoding follows each model's evaluation configuration: greedy
for Gemma-2 and Llama; temperature $0.6$, top-$p$ $0.95$, top-$k$ $20$ with
thinking disabled for Qwen3. Each response is then labelled as
\emph{obeyed} or \emph{refused} by Grok~4.7 used as an LLM judge, and the
obedience rate reported in Figure~\ref{fig:radars}(\captionc) is the fraction
of the 100 prompts labelled obeyed.

\section{AxBench Concept Datasets}
\label{app:axbench_dataset}

To evaluate the effectiveness, generalization, and transferability of steering methods, we extensively utilize the concept datasets introduced by \citep{wu2025axbench}. These datasets were developed as part of the AxBench benchmark and represent the largest publicly available collections of concept-conditioned instruction--response pairs for supervised dictionary learning (SDL) in LLMs. The datasets are constructed from concept features discovered in GemmaScope/LlamaScope and provide standardized training and evaluation data for thousands of interpretable concepts.

All datasets follow a common structure. Each example consists of:

\begin{itemize}
    \item \textbf{input}: An instruction sampled from publicly available instruction-tuning datasets spanning three domains: \textit{text}, \textit{code}, and \textit{mathematics}.
    \item \textbf{output}: A model-generated response. For positive examples, the response explicitly contains a target concept; for negative examples, the response does not contain any target concept.
    \item \textbf{output\_concept}: The concept expressed in the response. A special token (\texttt{EEEEE}) indicates the absence of a target concept.
    \item \textbf{concept\_genre}: The domain of the example (\textit{text}, \textit{code}, or \textit{math}).
    \item \textbf{category}: Indicates whether the example is a positive or negative instance of the target concept.
    \item \textbf{dataset\_category}: The dataset type. For all released SDL datasets, this field is set to \texttt{instruction}.
    \item \textbf{concept\_id}: A globally unique identifier corresponding to the discovered concept and its associated dictionary entry.
\end{itemize}

\subsection{Concept500}

\textbf{Concept500} is the primary evaluation dataset used in AxBench \cite{wu2025axbench}. It contains data for 500 concepts randomly sampled from the released GemmaScope concept inventory. The dataset includes concepts extracted from \texttt{\GemmaTwoTwoB} (layers 10 and 20) and \texttt{\GemmaTwoNineB} (layers 20 and 31).

For each concept, the dataset provides balanced positive and negative examples spanning text, code, and mathematical instructions. Each subset contains:

\begin{itemize}
    \item 720 positive examples (72 examples for each concept).
    \item 216 negative examples (72 examples per domain across three domains).
\end{itemize}

The relatively small scale of Concept500 makes it particularly suitable for controlled benchmarking and detailed analysis of steering performance across diverse concepts while maintaining manageable computational costs.

\subsection{Concept16K}

\textbf{Concept16K} significantly extends the scale of Concept500 and contains approximately 16,000 concepts randomly sampled from the GemmaScope concept inventory \cite{wu2025axbench}. Concepts are extracted from \texttt{\GemmaTwoTwoB} (layer 20) and \texttt{\GemmaTwoNineB} (layer 20).

At the time of its release, Concept16K represented the largest supervised dictionary learning dataset available for LLMs. The dataset was designed to support large-scale concept steering, representation learning, and evaluation studies. Despite its scale, the data generation pipeline remains computationally efficient, with the authors reporting a construction cost of less than \$0.01 per concept for generating 72 positive and 72 negative examples.

Compared to Concept500, Concept16K substantially increases concept diversity and enables evaluation of steering methods under a much broader distribution of semantic, stylistic, and functional concepts.

\subsection{Concept16K-v2}

\textbf{Concept16K-v2} extends Concept16K by incorporating concept data extracted from the \texttt{\LlamaThreeOneEightB} model family \cite{wu2025axbench}. The dataset preserves the same overall structure and concept coverage while improving response quality and diversity through longer generated outputs.

A key modification in Concept16K-v2 is the increase in average output sequence length from 64 tokens to 128 tokens. The longer generations provide richer contextual realizations of concepts and facilitate more comprehensive evaluation of concept steering methods, particularly for tasks that require sustained concept expression over extended outputs.

Consequently, Concept16K-v2 serves as a valuable benchmark for studying cross-model transferability, robustness, and scalability of steering approaches beyond the Gemma model family.

\subsection{LoRA Parameterization}
\label{app:lora_details}

All six backbones use an identical LoRA configuration, applied to every
transformer layer's four attention projections
($f\in\{q,k,v,o\}$, cf.\ Eq.~\ref{eq:lora-adapters}):

\begin{itemize}
    \item \textbf{Rank:} $r=16$
    \item \textbf{Scaling factor:} $\alpha=16$ (giving a fixed
    $\lambda_f=\alpha/r=1.0$ for all projections; $\lambda_f$ is a static
    hyperparameter, not optimized jointly with $\mathbf{A}_f,\mathbf{B}_f$
    during DPO training)
    \item \textbf{Dropout:} $0.1$, applied to the adapter input during
    training
    \item \textbf{Initialization:} standard LoRA initialization
    ($\mathbf{A}_f$ drawn from a Kaiming-uniform distribution,
    $\mathbf{B}_f$ initialized to zero), so that $\Delta\mathbf{W}_f=\mathbf{0}$
    and $\widetilde{\mathbf{W}}_f=\mathbf{W}_f$ at the start of training
    \item \textbf{Target modules:} \texttt{q\_proj}, \texttt{k\_proj},
    \texttt{v\_proj}, \texttt{o\_proj}
    \item \textbf{Layer coverage:} adapters are inserted at
    \emph{every} transformer layer of the backbone (no layer subsetting),
    yielding $4L$ adapted matrices per model, where $L$ is the backbone's
    layer count (Table~\ref{tab:lora-per-model})
\end{itemize}

\begin{table}[h]
\centering
\caption{Per-backbone LoRA configuration. Rank, $\alpha$, dropout, and
target modules are identical across all six models; only layer count
$L$ (and hence the number of adapted matrices $4L$), max sequence length,
and preference dataset version vary.}
\label{tab:lora-per-model}
\begin{tabular}{lccccc}
\toprule
Model & $L$ (layers) & Adapted matrices ($4L$) & Max seq.\ len.\ & Dataset \\
\midrule
Qwen3-1.7B            & 28 & 112 & 512 & Concept16K \\
Qwen3-4B               & 36 & 144 & 512 & Concept16K \\
Llama-3.2-3B-Instruct  & 28 & 112 & 256 & Concept16K \\
Llama-3.1-8B-Instruct  & 32 & 128 & 256 & Concept16K \\
Gemma-2-9B-it          & 42 & 168 & 256 & Concept16K-v2 \\
Gemma-2-2B-it          & 26 & 104 & 256 & Concept16K-v2 \\
\bottomrule
\end{tabular}
\end{table}

All other DPO hyperparameters (learning rate, warmup, effective batch size,
$\beta$) are shared across backbones and listed in
Appendix~\ref{app:dpo_details}.

\subsection{Preference Data Construction, Splits, and Leakage Control}
\label{app:dpo_dataset_construction}

\paragraph{Source data and pair construction.}
Training tuples $(x,c,y^+,y^-)$ (Eq.~\ref{eq:tuples}) are derived from the
AxBench \textsc{Concept16K} generation sets~\citep{wu2025axbench}:
\textsc{v1} (Gemma-2-2B/9B layer-20 SAE concepts, \texttt{text} genre only,
first ten concept identifiers removed) for Qwen3 and Llama, and \textsc{v2}
(Llama-3.1-8B layer-20, 131k-width SAE concepts, all genres) for Gemma-2. Each
AxBench row pairs an \texttt{input} with an \texttt{output} that either
exhibits \texttt{output\_concept} (\texttt{positive}) or is concept-free
(\texttt{negative}). We set $x=$\,\texttt{input} and $c=$\,\texttt{output\_concept}
(a natural-language SAE feature description, e.g.\ \emph{``Boolean indicators
of truth values''}). Instructions without a \texttt{negative} response are
discarded; for every remaining $(x,c)$ group with at least one
\texttt{positive} response we emit exactly one tuple, with $y^+$ the first
positive response and $y^-$ a negative response of $x$ drawn with a fixed seed
(\texttt{random\_state}=44). Since the negative pool and seed are identical for
all concepts sharing $x$, $y^-$ is the same concept-free response across those
concepts; $\mathcal{D}_c^-$ is thus concept-agnostic in practice and the
preference signal is carried by $y^+$. Tuples whose prompt, $y^+$, or $y^-$ is
shorter than 10 characters are dropped (61 in v1, 4 in v2). No further
filtering, deduplication, balancing, or subsampling is applied; each
instruction appears with many concepts ($\approx1{,}550$ on average in v1).

\paragraph{Prompt and concept formatting.}
Instruction and concept form a single user message via the fixed template
\begin{quote}\ttfamily
\#\#\# Instruction\textbackslash n\{$x$\}\textbackslash n\textbackslash n\#\#\# Steering Direction\textbackslash n\{$c$\}
\end{quote}
which is stored as the sole \texttt{user} turn of a conversational preference
record whose \texttt{chosen}/\texttt{rejected} fields hold $y^+$/$y^-$ as
single \texttt{assistant} turns; the model's native chat template is applied
by the trainer and no system prompt is used in training. The identical
template is used for all evaluations
(\S\ref{sec:zero-shots-text}, \S\ref{sec:zero-shots-agent}), with $c$ set to
the AxBench concept description, ``\texttt{Respond in \{Language\}.}'' for
CLaS-Bench, or a Big-Five trait description for PersonalityBench. At inference
the system prompt ``You are a helpful assistant.'' is prepended (folded into
the user turn for Gemma-2, whose template has no system role).

\paragraph{Split protocol.}
The unit of splitting is the \emph{concept}. The set $\mathcal{C}$ of concepts
with at least one tuple is shuffled with \texttt{numpy.random.default\_rng(42)}
and partitioned 75\,/\,10\,/\,15 into train/validation/test; all tuples of a
concept follow that concept, so the three concept sets are pairwise disjoint
(asserted programmatically). Instructions are \emph{not} held out: tuples
exist only for the 142 (v1) / 216 (v2) instructions with a concept-free
reference response, and every one of them appears in all splits paired with
different concepts. The held-out DPO test split therefore measures transfer
to \emph{unseen concepts on seen instructions}; generalisation to unseen
instructions is assessed only on the external benchmarks. Because one tuple is
emitted per $(x,c)$, pairs per concept equals instructions per concept.
Table~\ref{tab:axbench_splits} reports the statistics; $\mathcal{D}$ in
Eq.~\ref{eq:data-pool} is the train split (164{,}845 pairs for v1; 61{,}857
for v2).

\begin{table}[t]
\centering
\small
\setlength{\tabcolsep}{5pt}
\begin{tabular}{lrr}
\toprule
 & \textsc{Concept16K-v1} & \textsc{Concept16K-v2} \\
 & (Qwen3, Llama) & (Gemma-2) \\
\midrule
SAE dictionary (concept source) & Gemma-2-2B/9B, L20 & Llama-3.1-8B, L20 (131k) \\
Raw rows after filtering & 1{,}538{,}640 & 1{,}152{,}216 \\
Distinct instructions / with concept-free response & 996 / 142 & 2{,}995 / 216 \\
Tuples after length filter & 220{,}090 & 82{,}586 \\
Concepts with $\geq 1$ tuple, $|\mathcal{C}|$ & 21{,}028 & 15{,}706 \\
Pairs (= instructions) per concept: min / median / max & 2 / 10 / 99 & 1 / 5 / 70 \\
Distinct $y^-$ (one per instruction) & 142 & 216 \\
Mean length $y^+$ / $y^-$ (chars) & 352 / 652 & 536 / 703 \\
\midrule
Train: concepts / pairs & 15{,}771 / 164{,}845 & 11{,}779 / 61{,}857 \\
Validation: concepts / pairs & 2{,}102 / 22{,}127 & 1{,}570 / 8{,}282 \\
Test: concepts / pairs & 3{,}155 / 33{,}118 & 2{,}357 / 12{,}447 \\
Instructions per split & 142 (all shared) & 216 (all shared) \\
Optimizer steps (2 epochs, eff.\ batch 128) & 2{,}576 & 968 \\
\bottomrule
\end{tabular}
\caption{Construction and concept-level split statistics (seed 42) of the
pooled preference data $\mathcal{D}$. Splits are disjoint in concepts;
instructions are shared across splits by construction. Counts were
regenerated from the public AxBench parquet files with the released
preprocessing scripts and fixed seeds.}
\label{tab:axbench_splits}
\end{table}

\paragraph{Validation, hyperparameter selection, and leakage prevention.}
Validation and test concepts are disjoint from each other and from training
concepts, and test tuples are never used for training, model selection, early
stopping, or checkpoint selection. All hyperparameters were fixed a priori and
shared across models: learning rate $10^{-5}$ (linear schedule, 3\% warm-up),
weight decay $0.01$, DPO $\beta=0.1$ (sigmoid loss, no label smoothing), 2
epochs, effective batch 128 ($16\times8$ accumulation), bf16 with gradient
checkpointing, LoRA $r=16$, $\alpha=16$, dropout $0.1$, no bias, on
\texttt{q\_proj}, \texttt{k\_proj}, \texttt{v\_proj}, \texttt{o\_proj}; maximum
length 256 tokens (Gemma-2, Llama) or 512 (Qwen3); frozen base model as
reference policy. No validation-based or benchmark-specific tuning was
performed. A fixed 500-tuple validation subsample (seed 42) is evaluated every
200 steps solely to monitor DPO loss and reward accuracy. Checkpoints are
saved every 400 steps with \texttt{save\_total\_limit}=1, so only the final
checkpoint (step 2{,}576 for v1, 968 for v2; end of epoch~2) exists and is
evaluated on all benchmarks, precluding checkpoint selection. A 500-tuple test
subsample is scored once after training as an in-distribution sanity check.
Training seed 42 throughout; TRL 1.8.0, PEFT 0.19.1, Transformers 5.14.1, a
single GPU (CUDA 13.2).

\end{document}